\documentclass{ieeetj}
\usepackage{amsmath,amssymb,amsfonts}
\usepackage{algorithmic}
\usepackage{graphicx,color}
\usepackage{textcomp}
\usepackage{hyperref}
\usepackage{multirow}
\usepackage{tabularx}
\hypersetup{
  colorlinks=true,    
  citecolor=blue,     
  linkcolor=blue,    
  urlcolor=black      
}
\usepackage{algorithm,algorithmic}
\usepackage{physics}
\def\BibTeX{{\rm B\kern-.05em{\sc i\kern-.025em b}\kern-.08em
    T\kern-.1667em\lower.7ex\hbox{E}\kern-.125emX}}
\AtBeginDocument{\definecolor{tmlcncolor}{cmyk}{0.93,0.59,0.15,0.02}\definecolor{NavyBlue}{RGB}{0,86,125}}

\usepackage[l2tabu,orthodox]{nag}

\usepackage[
backend=biber, 
bibencoding=utf8,
style=ieee, 
sorting=none, 
natbib=true, 
doi=false, 
isbn=false, 
url=false, 
eprint=false, 
maxcitenames=1, 
mincitenames=1,
maxbibnames=5,
]{biblatex}
\AtEveryBibitem{\clearlist{language}}

\usepackage[printonlyused]{acronym}
    \acrodef{GNSS}{Global Navigation Satellite System}
    \acrodef{RTK}{Real Time Kinematic}
    \acrodef{ECU}{Electronic Control Unit}
    \acrodef{MPC}{Model Predictive Control}
    \acrodef{NMPC}{Nonlinear Model Predictive Control}
    \acrodef{PID}{Proportional-Integral-Derivative}
    \acrodef{IMU}{Inertial Motion Unit}
    \acrodef{ANR}{French National Research Agency}
    \acrodef{TIARA}{Toward Intelligent Adaptable Robots for Agriculture}
    \acrodef{TSCF}{Technologies et systèmes d’information pour les agrosystèmes-Clermont-Ferrand}
    \acrodef{LIDAR}{LIght Detection And Ranging}
    \acrodef{DARPA}{Defense Advanced Research Projects Agency}
    \acrodef{RMSE}{Root Mean Square Error}
    \acrodef{FAO}{Food and Agriculture Organization of the United Nations}
    \acrodef{IQR}{InterQuartile Range}
    \acrodef{ROS}{Robot Operating System}

\usepackage{siunitx}
\usepackage[all]{nowidow}

\usepackage[dvipsnames]{xcolor}

\usepackage{lipsum}

\usepackage{subcaption}

\usepackage{epstopdf}

\usepackage{import}

\graphicspath{{./latexGoodPractices/}}

\usepackage{booktabs}

\usepackage{tabu}

\usepackage{amssymb,amsfonts,amsmath,amscd}

\usepackage{bm} 

\usepackage{stfloats}

\newcommand{\bbm}{\begin{bmatrix}}
\newcommand{\ebm}{\end{bmatrix}}

\usepackage{xspace}
\newcommand{\eg}{e.g.,\xspace}
\newcommand{\ie}{i.e.,\xspace}

\newcommand{\review}[1]{\textcolor{black}{#1}}
\newcommand{\rereview}[1]{\textcolor{black}{#1}}
\newcommand{\defensereview}[1]{\textcolor{black}{#1}}

\def\OJlogo{\vspace{-4pt}$<$Society logo(s) and publication title will appear here.$>$}
\def\seclogo{\vspace{10pt}$<$Society logo(s) and publication title will appear here.$>$}

\def\authorrefmark#1{\ensuremath{^{\textbf{#1}}}}

\begin{document}
\receiveddate{XX Month, XXXX}
\reviseddate{XX Month, XXXX}
\accepteddate{XX Month, XXXX}
\publisheddate{XX Month, XXXX}
\currentdate{XX Month, XXXX}
\doiinfo{XXXX.2024.123456789}

\markboth{A Predictive Control Strategy to Offset-Point Tracking for Agricultural Mobile Robots}{Author {et al.}}

\title{A Predictive Control Strategy to Offset-Point Tracking for Agricultural Mobile Robots}

\author{Stephane Ngnepiepaye Wembe\authorrefmark{1,2}, Vincent Rousseau\authorrefmark{1}, Johann Laconte\authorrefmark{1} and Roland Lenain\authorrefmark{1}}
\affil{
Université Clermont Auvergne, INRAE, UR TSCF, 63000, Clermont-Ferrand, France; stephane.ngnepiepaye-wembe@inrae.fr
}
\affil{
SABI AGRI, 63360, Saint-Beauzire, France; stephane.ngnepiepaye-wembe@sabi-agri.com
}
\corresp{
Corresponding author: 
Stephane Ngnepiepaye Wembe (email: sngnepiepayewembe@gmail.com)
}
\authornote{
This work has been funded by the french National Research Agency (ANR), under the grant ANR-19-LCV2-0011, attributed to the joint laboratory Tiara (\url{www6.inrae.fr/tiara}). It has also received the support of the French government research program "Investissements d'Avenir" through the IDEX-ISITE initiative 16-IDEX-0001 (CAP 20-25), the IMobS3 Laboratory of Excellence (ANR-10-LABX-16-01).  This work has been partially supported by ROBOTEX 2.0 (Grants ROBOTEX ANR-10-EQPX-44-01 and TIRREX ANR-21-ESRE-0015)
}

\begin{abstract}
\review{Robots are increasingly being deployed in agriculture to support sustainable practices and improve productivity. They offer strong potential to enable precise, efficient, and environmentally friendly operations. However, most existing path-following controllers focus solely on the robot’s center of motion and neglect the spatial footprint and dynamics of attached implements. In practice, implements such as mechanical weeders or spring-tine cultivators are often large, rigidly mounted, and directly interacting with crops and soil; ignoring their position can degrade tracking performance and increase the risk of crop damage.
To address this limitation, we propose a \rereview{closed-form} predictive control strategy extending the approach introduced in \cite{offset_point_control_article}.
The method is developed specifically for Ackermann-type agricultural vehicles and explicitly models the implement as a rigid offset point, while accounting for lateral slip and lever-arm effects.
The approach is benchmarked against state-of-the-art baseline controllers, including a reactive geometric method, a reactive backstepping method, and a model-based predictive scheme. \rereview{Real-world agricultural experiments with two different implements show that the proposed method reduces the median tracking error by $\SI{24}{\%}$ to $\SI{56}{\%}$, and decreases peak errors during curvature transitions by up to $\SI{70}{\%}$.} These improvements translate into enhanced operational safety, particularly in scenarios where the implement operates in close proximity to crop rows.}
\end{abstract}



\maketitle
\section{INTRODUCTION}
\IEEEPARstart{A}{griculture} as a whole faces a multitude of challenges that arise, directly or indirectly, from the realities of the modern world, such as global population growth or pollution.
Among the most pressing issues are environmental constraints, such as soil preservation, and the imperative to ensure sufficient production to meet the ever-increasing global food demand. 
This last challenge is underscored by recent data from the~\ac{FAO}, which report that approximately 733 million people suffered from hunger in 2023, representing one in eleven individuals globally~\cite{fao_state_2024}.
In response to this challenge, two main approaches have emerged in agriculture.
One approach prioritizes environmentally friendly farming practices, such as conservation agriculture. 
By promoting sustainable soil management and reducing chemical inputs, it helps limit ecological degradation. 
However, this method often results in relatively modest yields that fall short of meeting the growing global food demand.
In contrast, a second approach, widely adopted worldwide, relies on intensive agriculture. 
It focuses on maximizing production through the extensive use of pesticides, chemical fertilizers, and other potentially harmful substances. 
While this strategy enables large-scale output, it also causes significant environmental harm, threatening the long-term sustainability of natural resources.
Reconciling these two fundamental objectives, \ie high productivity and environmental preservation, has long posed a challenge. 
This involves strategic trade-offs, forcing stakeholders to choose between prioritizing agricultural output and safeguarding ecosystems.

To address this dual challenge of ensuring consistent agricultural production while respecting the environment, a new approach has emerged: agroecology.
Today, this vision has become the guiding principle of sustainable agriculture. 
Agroecology seeks to achieve sufficient production levels while incorporating practices that respect ecological constraints, such as soil preservation and minimizing environmental impact
\cite{tilman_agricultural_2002}.
In this context, Agriculture 4.0 plays a pivotal role, integrating advanced technologies such as robotics and automated equipment to support sustainable and environmentally friendly farming~\cite{santos_valle_agriculture_2020}. 
Agricultural robotics, in particular, has emerged as a promising technological solution~\cite{lenain2021agricultural}. 
It aims to transform the agricultural sector by automating and optimizing key tasks, thereby addressing pressing challenges such as labor shortages and the demand for more sustainable agricultural practices~\cite{lenain_agricultural_2019}.
Robotic systems are increasingly becoming capable of performing repetitive operations, such as planting, weeding, spraying, and harvesting, with high precision. 
These innovations contribute to the reduction of pesticide use, thereby mitigating environmental impact. 
Moreover, they alleviate the physical burden on workers, supporting the transition toward more agroecological and sustainable farming systems.

The growing interest in agricultural robotics is reflected in a multitude of research projects aimed at developing high-performance robotic systems for field operations.
These projects aim to improve both productivity and sustainability in the agricultural sector~\cite{inbook_Robotics_in_Agriculture}. 
To perform an agricultural task, such as weeding, a robot must be equipped with several essential functions~\cite{Software_Architecture_Muzaffar}. 
In agricultural applications, the tasks assigned to robots, whether sowing, weeding, or harvesting, share a fundamental objective: precise path-following. 
This capability is critical to ensure effective interventions. 

\begin{figure}[t]
  \centering
  \includegraphics[width=\linewidth]{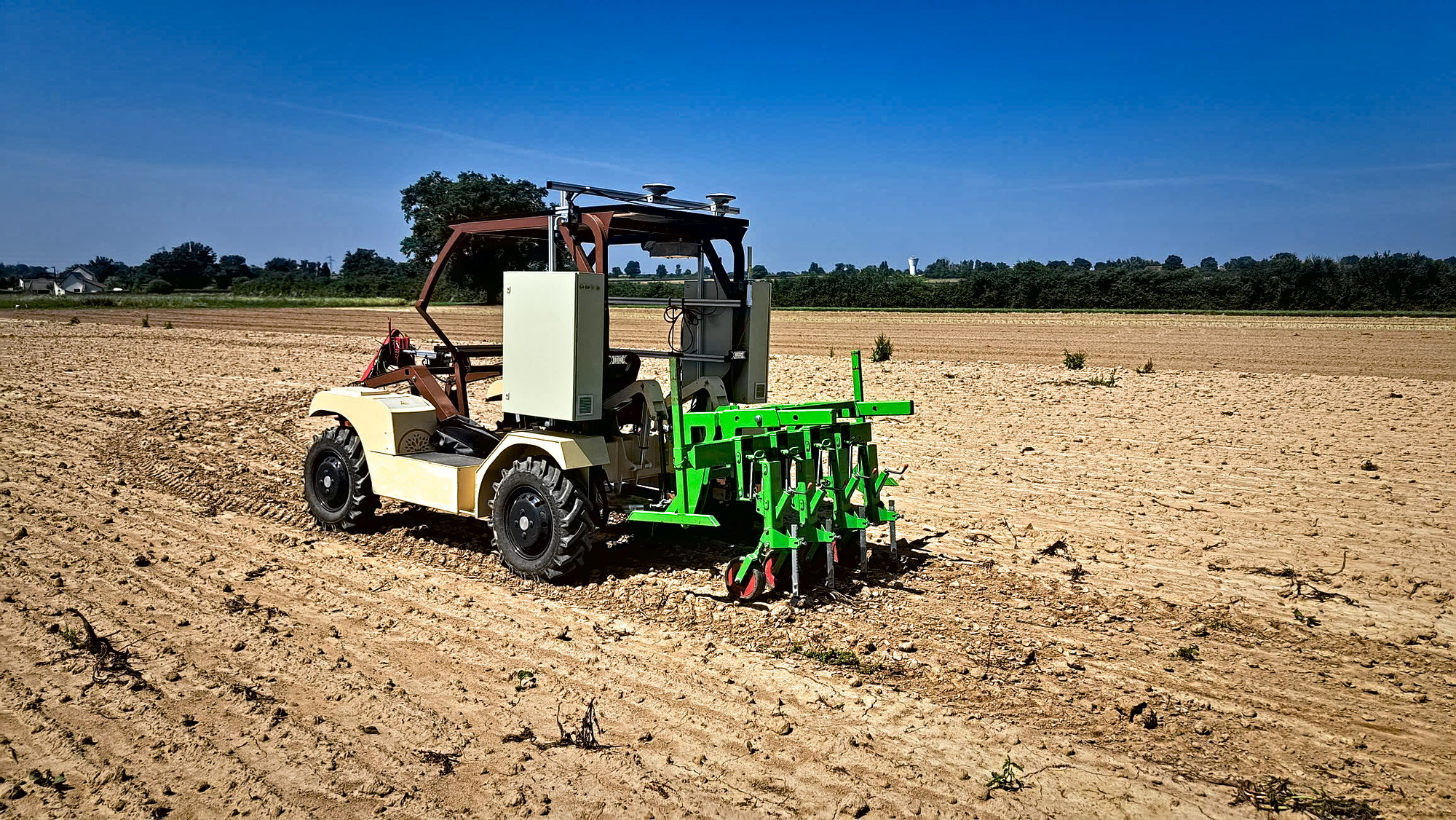}
  \caption{An autonomous agricultural system, consisting of a robot linked to a mounted implement. In this configuration, the control accuracy is required at the implement-soil interaction point rather than at the center of the robot, which renders center-based control laws inadequate.}
  \label{fig:intro}
\end{figure}

During an agricultural operation, the implement attached to the robot must often interact directly with crops.
In weeding operations, the end effector, such as a blade or plowshare, must follow a precise path that allows it to remove weeds while preserving crops.
\autoref{fig:intro} depicts a robot performing an inter-row weeding operation. 
In this scenario, the effector elements are the plowshares, which carry out the weeding task. 
These plowshares follow a designated path, allowing them to operate near the crop rows. 
However, if the path-following algorithm is ineffective, high magnitude oscillations may occur, potentially causing the effector elements of the agricultural implement to damage the crops. 
An optimized path-following controller ensures not only the efficiency of the operation but also minimizes the risk of damage to crops.
Traditional control strategies typically focus on the rear axle center of the robot. 
This choice has the advantage of allowing linearizable modeling using exact linearization techniques~\cite{samson2016modeling}. 
However, as illustrated in~\autoref{fig:intro}, agricultural applications introduce a specific constraint: it is the implement, not the robot’s center, that directly interacts with the soil and crops.
As a result, the control algorithm must be specifically designed to account for the position and actions of the agricultural implement. 
This necessitates shifting the focus of control from the rear axle center to the agricultural implement itself to ensure maximum precision in performing agricultural tasks, such as weeding or sowing, where direct contact with crops is critical.
Controlling the agricultural implement directly, rather than the rear axle center, involves managing an offset point of the system while maintaining the vehicle’s overall kinematics. 
As such, this requires the development of specific control laws that enable the offset point to follow a predefined path accurately.

In the literature, the issue of controlling an offset point relative to the robot is related to the control of trailer systems.
However, trailer systems have the advantage of a pivot connection at the attachment point: during curvature, the pivot connection contributes to reducing the tracking error by rotating. 
Furthermore, trailer systems are generally limited to configurations where the trailer is positioned at the rear of the vehicle.
Nevertheless, in many agricultural applications, such as the use of inter-vine blades in vineyards or weeders in large-scale farming, agricultural implements are often mounted at the front of the vehicle and connected to the vehicle via a rigid link. 
This configuration introduces additional challenges: the implements are bulky, and the rigid connection creates a significant lever arm, which can severely degrade the precision of the operation.
This loss of precision can have serious consequences, particularly in delicate operations such as weeding, where irreversible damage to the crops may occur~\cite{gan2007implement}.

In this paper, we extend the control strategy introduced in~\cite{offset_point_control_article}, originally designed to regulate an offset point relative to the rear axle center. The extension explicitly accounts for lateral sliding effects and introduces a predictive control approach aimed at reducing tracking errors, which can be unavoidable. A key challenge in this context is indeed that the reference trajectories are inherently infeasible: the same path must often be executed with implements of varying sizes and lever arms, making perfect tracking unattainable. The role of the proposed method is therefore not to suppress tracking errors entirely, but to minimize them as effectively as possible. We validate the approach through real-world agricultural experiments, highlighting the advantages of predictive control for implements rigidly attached to the robotic platform, a configuration commonly encountered in practice.

\review{
The main contributions of this work can be summarized as follows:
\begin{itemize}
    \item An extension of the model introduced in~\cite{offset_point_control_article} that explicitly accounts for lateral sliding effects; and 
    \item A novel predictive control strategy for rigidly linked implements.
\end{itemize}
}


\rereview{
\autoref{sec:rw} reviews the state of the art in path-following control, with emphasis on agricultural implements and trailer-like systems. 
\autoref{sec:theory} extends the model introduced in \cite{offset_point_control_article} by incorporating lateral sliding effects and proposes a novel control strategy for rigidly linked implements. 
The proposed developments are validated in simulation in \autoref{sec:simulation}, where the influence of key system parameters is systematically analyzed. 
\autoref{sec:exp} provides a quantitative comparison with state-of-the-art approaches and reports real-world experimental results in an agricultural scenario, demonstrating the effectiveness of the proposed method. 
Finally, \autoref{sec:discussion} discusses the main insights derived from this work and outlines future research directions.}

\section{RELATED WORKS}\label{sec:rw}
In this section, we present an overview of the path-following approaches applied to the agricultural domain. 
We describe the most recent methods designed to ensure that a robot follows a path precisely, with a particular focus on control strategies for agricultural implements.

\subsection{Path Following Focused on the Center of the Robot} 
Several approaches have been developed to address the challenge of keeping a robot’s center on a path, depending on specific needs. Three main approaches are commonly distinguished.
The first relies on point-to-point motion, where the robot progresses from a known starting point to a predefined endpoint~\cite{article_point_to_point}. 
The second involves trajectory tracking, which consists of following a path while adhering to precise temporal constraints~\cite{inproceedings_Time_constrained_trajectory_tracking}. 
The third approach, path-following, differs in that it has no temporal constraints~\cite{path_following_based_on_position}.
In the agricultural domain, path-following is generally favored. 
This method ensures precise path tracking while enabling the robot to adapt its behavior to specific terrain conditions and the operational requirements of agricultural tasks. 
Since the path-following process is not time-constrained, the robot can allocate additional time based on the task complexity.

Path-following algorithms can be categorized into two groups: geometric-based control methods and model-based methods.

\subsubsection{Geometric-Based Control Methods}
Geometric-based control methods rely on an analysis of the geometric situation to define the control laws necessary for the center of the robot to follow a predefined path. 
In this category, the most popular approaches are Pure Pursuit and the Stanley controller. 

Originally designed to guide missiles toward moving targets~\cite{Scharf_pure_pursuit}, the Pure Pursuit controller was later adapted for path-following~\cite{Coulter-1992-13338}. 
This method involves determining, at each moment, a target point on the reference path that the robot must reach. 
The target point is identified using a parameter called the 'look-ahead distance', which represents the distance at which the target point is positioned ahead of the robot.
The applied control ensures that the robot follows the circular arc connecting its center to the target point on the reference path.
Due to its simplicity, this algorithm has been widely used. 
However, one of the main challenges of the Pure Pursuit controller lies in selecting the 'look-ahead distance' parameter. 
An improper adjustment of this parameter can lead to undesirable control oscillations.
To overcome this limitation, several adaptations of the algorithm have been developed, leading to new variants widely used today in the agricultural domain. 
An improvement in which the 'look-ahead distance' parameter is automatically adjusted based on the angular deviation is proposed in~\cite{Qiang_pure_pursuit_agricole}, enabling a tractor to follow a path with greater precision. 
 A method that identifies a target point, minimizing both lateral and angular errors over a distance horizon defined by the 'look-ahead distance' is introduced in ~\cite{yang_optimal_2022}.
Despite these advancements, the Pure Pursuit controller exhibits inherent design limitations. In particular, its control computation does not consider the vehicle’s orientation, which can significantly degrade performance in the presence of large angular errors~\cite{rokonuzzaman_review_2021} and thus may cause damage to crops in agricultural environments.

The Stanley controller, also known as the Hoffmann controller, was first implemented by the winners of the~\ac{DARPA} Grand Challenge in 2005~\cite{thrun_stanley_2007}.
While conceptually similar to Pure Pursuit, it differs in its approach: the orientation of the vehicle relative to the reference path and the steering angle are directly integrated into the steering control calculation. 
This key distinction eliminates the need for the 'look-ahead distance' parameter~\cite{Hoffmann_standley}.
However, the Stanley controller is focused on the current tracking error and not future path segments and is therefore prone to oscillation~\cite{ms-12-419-2021}, especially when following curved paths. 

Controllers based on geometric analysis allow for the development of control laws that are relatively easy to set up. 
However, relying solely on geometric parameters can impact the smoothness of movement, leading to abrupt motions that may be hazardous in agricultural operations~\cite{rokonuzzaman_review_2021}.

\subsubsection{Model-Based Methods}
Unlike geometric methods, Model-based methods rely on a system model, typically kinematic, to provide a more precise description of its motion.
Kinematic-based methods enable the design of control laws expressed in terms of velocities and accelerations, promoting smoother and more precise trajectories~\cite{rokonuzzaman_review_2021}. 
This category encompasses various algorithms, including feedback controllers, predictive approaches, and adaptive methods.

Feedback control aims to minimize the error between the reference signal of the system and its output.
This is usually done with a linear relationship between the control input and the output, thereby facilitating the application of classical control methods, such as~\ac{PID} controllers, to generate the system inputs.
The specificity of the methods proposed in the literature lies in the choice of the linearization technique used, each approach offering advantages tailored to different contexts.
Linearization can be achieved using the state-feedback transformation. The authors of~\cite{Samson_Feedback} linearize the system around an operating point to design the control law. 
However, this approach has a significant limitation: the control law becomes invalid if the system deviates from this operating point.
An alternative involves performing exact linearization by exploiting a chained form of the system, as proposed in~\cite{Samson_chained}. 
This method enables global linearization, provided that the model of the system is compatible with this structure.
However, it is important to note that linearization is not always applicable, especially for non-linear systems with singularities~\cite{rokonuzzaman_review_2021}. 
These constraints have driven the development of adaptive methods that offer a more flexible and robust solution to the limitations of linearization approaches.

Among adaptive control methods, backstepping occupies a prominent place.
Originally developed in the 1990s in~\cite{Kokotovic_backstepping}, this non-linear control strategy aims to stabilize complex dynamic systems by recursively decomposing them into simpler subsystems.
For each subsystem, a stabilizing control law is derived using Lyapunov theory.
Since the output of one subsystem serves as the input to the next, this recursive design process ensures the overall stability of the full non-linear system.
Although initially developed for stabilization tasks, backstepping has since been successfully extended to path-following applications~\cite{Fukao_adaptive_backstepping}.
In such contexts, the control objective involves stabilizing variables such as the vehicle’s orientation and tracking error~\cite{Imen_Backstepping}. However, other internal or intermediate variables, such as articulation angles, can also be included in the design process to improve stability and tracking performance.
In~\cite{Huynh_backstepping}, a model of a tractor towing a pivotable trailer is presented. 
Backstepping control is then applied to ensure precise trajectory tracking, including stabilization of the trailer’s articulation angle.
Despite its advantages of stabilizing intermediate variables, backstepping involves inherent complexity in selecting appropriate subsystem structures and tuning control gains, particularly for high-order or strongly coupled systems. 
Similarly to linearization approaches, backstepping relies on an accurate system model, making it sensitive to uncertainties and external disturbances. Moreover, it considers current errors only and does not account for actuator settling times or yaw-rate servo delays. To enhance robustness against delays, an alternative is to adopt a predictive approach, which anticipates potential changes in the system or its environment.

One of the most widespread predictive approaches is the \ac{MPC}. 
This method estimates the future state of the system over a given prediction horizon. 
Based on this estimation, an objective function to be minimized is defined, subject to specific constraints. 
The controller then determines the control sequence to apply over the prediction horizon to achieve the minimization objective. 
Only the first control action of this sequence is executed by the robot, and the process is repeated at each control cycle.
~\ac{MPC} is widely used for its performance, and numerous enhancements to this method have led to various variants, as reviewed in~\cite{ding_model_2018}. 
In~\cite{backman_collision_2013}, a cost function is designed to simultaneously minimize tracking errors and control effort while maximizing the distance to obstacles. However, the algorithm presented in that study is computationally more intensive than baseline methods due to the use of a numerical solver. The reliability of this approach is also limited, as the solver may occasionally fail to find a control input, causing the system to retain the previous command~\cite{backman_navigation_2012}.

Model-based approaches enable the design of control laws that ensure smooth and stable motion. Backstepping is a representative example of such methods, as it allows for the control of intermediate variables through a recursive design. However, its reactive nature makes it sensitive to external disturbances. In contrast, predictive approaches such as~\ac{MPC} anticipating future system behavior and external influences, thereby achieving higher precision. Nonetheless, these methods are often computationally intensive and can suffer from reliability issues, particularly when the solver fails to find a feasible solution. \review{In this paper, we propose a predictive closed-form approach that retains the high predictive performance of an~\ac{MPC} while guaranteeing convergence, thus improving the overall reliability of the control strategy.
}

\subsection{Control of Agricultural Implements}
In essence, robots used in large agricultural operations primarily serve as implement carriers. 
The attached implements perform the agricultural tasks, as they are directly in contact with the soil and plants. 
Therefore, the focus of the control algorithm should be on the implement itself, rather than the robot. 

The classification of implement control algorithms can be based on the nature of the connection between the implement and the carrier (\ie the robot). 
This connection can be a pivot type, corresponding to trailer systems, or a rigid type, which is typically found in modern implement-carrying agricultural robots.

\subsubsection{Control of Trailer Systems}
A trailer system corresponds to a vehicle equipped with an additional structure located at the rear and linked to the robot via a pivot link.
The specificity of these systems lies in the construction of their model, which can include additional state variables such as the pivot angle between the carrier and the implement~\cite{Cariou_trailer}. 
The control laws, once a suitable model is established, are similar to those previously discussed. An \ac{MPC} approach to the model of a trailer system, incorporating the pivot angle as a state variable to achieve precise path-following, is applied in \cite{Wu_trailer_mpc}. 
Similarly, the control of an active implement mounted on a tractor via a controllable pivot joint is proposed in \cite{backman_nonlinear_2010}. 
As in the previous case, the state variables include the pivot angle, which can be controlled using an actuator. 
A~\ac{NMPC} method is then used to design a control system that integrates both the front-wheel steering angle of the carrier and the pivot angle between the implement and the tractor.
This dual control, involving both the tractor and the pivot joint, achieves tracking precision within $\SI{8}{\cm}$, even at speeds of up to $\SI{12}{km/h}$. 
Whether controlled or not, the pivot joint between the tractor and the implement plays a crucial role in reducing tracking errors, especially during turns.

\subsubsection{Control of Implements Connected via a Rigid Joint} 
Agricultural robots are predominantly utilized as carriers for various implements. 
The specific implement is chosen based on the agricultural task to be executed (sowing or weeding) and subsequently mounted onto the robotic tractor.
In most cases, these implements are rigidly attached to the tractor, without a pivot joint. 
Depending on the width of the implement and its longitudinal offset, the lever arm between the center of the rear axle of the tractor and the working point of the implement can be significant, leading to substantial tracking errors, particularly in curvatures~\cite{gan2007implement}.
To correct these tracking errors, the use of an automatic hitch capable of compensating for the tracking errors of the implement is proposed in~\cite{Semichev_article_hitch}. Similarly, integrating an~\ac{ECU} that considers the implement and the tractor as two distinct systems is suggested in~\cite{Freimann2007ABA}. 
However, this solution assumes that the robot is equipped with an active implement, which is not always the case.
While treating the implement and tractor as decoupled systems simplifies the model, it introduces limitations, most notably the omission of mutual interactions, which can adversely affect tracking performance.

To the best of our knowledge, only~\cite{lukassek_model_2020} and~\cite{offset_point_control_article} have addressed the problem of trajectory tracking for an offset point on an agricultural implement rigidly attached to a robot.
In~\cite{offset_point_control_article}, two control strategies are proposed. The first relies on lateral servoing at the robot’s center, while the second employs a backstepping-based control law on the implement. Both methods enable the offset point to track a desired path; however, they are reactive in nature and exhibit sensitivity to external disturbances, particularly during changes in curvature.
In contrast, a predictive control strategy, reformulating the system model on the implement point to be controlled, is introduced in~\cite{lukassek_model_2020}.
However, their method is confined to configurations where the implement is mounted at the front of the tractor, and no real-world experimental validation has been reported. Additionally, a numerical solver is necessary for solving the proposed~\ac{MPC} formulation, which can be computationally demanding and less reliable~\cite{backman_navigation_2012}.

In this paper, we compare our method to those in \cite{offset_point_control_article, lukassek_model_2020} in real-world scenarios, and show that the method in \cite{lukassek_model_2020} is not applicable to the case of a rear-mounted implement. In contrast, our method is applicable to both cases (front and rear) while having a closed form that discards the need for numerical solvers.


\section{APPROACH}\label{sec:theory}
In this section, we present a novel \ac{MPC} approach for a rigidly linked implement, either on the front or back of the robot.
First, we present the notations, hypotheses, and modeling of the robot.
Then, we extend the method in \cite{offset_point_control_article} to consider slippage.
Finally, we present our novel predictive method to control a rigidly linked implement at either the front or rear of the robot.

\subsection{Preliminary}
Consider an Ackermann-type robot as depicted in~\autoref{fig:intro} to which an implement is attached.  
The connection between the robot and the implement is considered rigid, preventing any movement of the implement relative to the robot during the motion of the robot. 
Let there $I$ be a point on this implement, and let there $\Pi_\text{ref}$ be the reference path.  
The problem consists of designing a control law for the vehicle, ensuring that the point $I$ follows the reference path $\Pi_\text{ref}$.
\subsubsection{Hypothesis}
As done in \cite{offset_point_control_article}, the following developments are made under the given assumptions:
\begin{description} 
    \item[H1] The wheels of the vehicle are assumed to remain in contact with the ground.
    \item[H2] Non-ideal grip conditions are characterized by the introduction of sideslip angles. 
    \item[H3] Dynamic effects other than sliding and skidding are negligible. 
    \item[H4] The offset point remains rigidly attached to the vehicle.
    \item[H5] The implement point’s offset does not exceed the trajectory’s minimum curvature radius. 
    \item[H6] The robot possesses a vertical sagittal plane of symmetry that intersects the center of the robot’s rear axle.
\end{description}

The first three hypotheses are valid when the vehicle operates in a natural environment, such as an agricultural one. 
Under these conditions, a kinematic model is sufficient to characterize the behavior of the robot, providing the estimation of sideslip angles. 
Hypothesis H4 corresponds to the typical linkage found in agricultural vehicles, where the implement is rigidly fixed to the vehicle without any pivoting connection. 
In practice, agricultural machines commonly use a three-point hitch, which provides this rigid attachment.
Hypothesis H5 is essential for the modeling described in the next sections, ensuring well-defined behaviors and avoiding undefined scenarios, as discussed in the subsequent section.
Finally, the last hypothesis, H6 is satisfied for any Ackermann-type vehicle. 
This symmetry simplifies the model of the robot, allowing it to be described as a bicycle-type vehicle.
\begin{figure}[h]
    \centering
    \includegraphics[width=\linewidth]{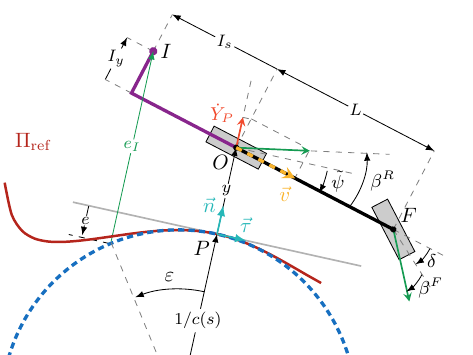}
    \caption{Notations used in this paper.
    Unlike standard robotics control problems, the objective is not to steer the robot’s center $O$ to converge to the reference path $\Pi_\text{ref}$ but rather to an offset point $I$ that represents the agricultural implement in this context. 
    The lateral perturbation $\dot Y_P$ arises from the rear sideslip angle  $\beta^R$, while the front sideslip angle $\beta^F$ is observed at the front wheel. \review{The tracking errors are computed in the Frenet frame $(\Vec{\tau}, \Vec{n})$. 
    The robot frame
    is such that $I_y$ is positive when the implement point $I$ is at the left, and $I_s$ is positive when the point $I$ is at the front. The theoretical velocity of the robot is denoted by $\Vec{v}$, and the green arrow represents the actual velocity due to the sideslip angles.}
    }
    \label{fig:model}
\end{figure}
\vspace{-0.5cm}
\subsubsection{Modeling}
The notations used in the robot modeling are illustrated in~\autoref{fig:model}.
As described above, the robot is modeled using a bicycle model in accordance with Hypothesis $H6$.
The offset point to be controlled is denoted by $I$ and may represent, for instance, the contact point between the implement and the ground.

The control point on the agricultural implement is defined by its local coordinates $(I_s, I_y)$, measured relative to the robot frame originating at the point $O$.
The value of these coordinates can be either positive \review{(to the left of the robot for $I_y$, in front of it for $I_s$)} or negative \review{(to the right of the robot for $I_y$, at the rear for $I_s$)}, allowing the implement to be positioned at the front, rear, left, and right of the center of the robot. 
Within this framework,  $\tilde{\psi}$ denotes the angular deviation and $y$ corresponds to the tracking error at the center of the rear axle (point $O$) relative to the reference path $\Pi_\text{ref}$.
Both $\tilde{\psi}$ and $y$  are measured using a~\ac{GNSS}-\ac{RTK} sensor and a geo-referenced path. 
The path $\Pi_\text{ref}$ is locally circular, with the curvature $c(s)$ defined at the curvilinear abscissa $s$.
Sliding effects are modeled by two sideslip angles, $\beta^R$ and $\beta^F$  corresponding to the rear and front wheels, respectively, as done in~\cite{sideslipRoland}. In this paper, we consider that the rear sideslip perturbation generates a lateral velocity disturbance denoted by $\dot{Y}_P$ applied to $O$.

This paper addresses the regulation of the implement tracking error $e_I$, defined as the error from the reference path $\Pi_\text{ref}$ at the point on the implement closest to the path, measured along a direction parallel to the distance $y$.
Thus, the tracking error of the offset point $I$ is defined as
\begin{equation}
	\begin{aligned}
		e_I &= y+ I_s \sin\tilde{\psi} + I_y\cos{\tilde{\psi}}+ e,
	\end{aligned}
    \label{eq:RelationEcart}
\end{equation}
with
\begin{equation}
	\begin{aligned}
		e &= -\frac{1}{c(s)} \left( 1- \cos\varepsilon \right), \quad\text{and}\\
		\varepsilon &= \arcsin \left(c(s)\left(I_s\cos\tilde{\psi}+ I_y\sin\tilde{\psi}\right)\right).
	\end{aligned}
\end{equation}

When the curvature $c(s)$ is zero, the distance $e$ is null, as easily seen from~\autoref{fig:model}.
Furthermore, as explained in~\cite{offset_point_control_article}, \autoref{eq:RelationEcart} exists if
\begin{equation}
\begin{aligned}
   \sqrt{I_s^2+I_y^2} < \left|\frac{1}{c(s)}\right|.
\end{aligned}
\label{eq:cond_TsTy}
\end{equation}
Accordingly, the distance of the control point $I$ from the robot's center must not exceed the radius of curvature, in accordance with the hypothesis $H5$. 
Note that in agricultural scenarios, high-curvature paths are not possible due to the maximum steering angle of the vehicles, and thus $H5$ is always satisfied in practical applications.

\subsubsection{Kinematic Model of the robot's center $O$}
The kinematics of the point $O$ are given by
\begin{equation}
    \begin{aligned}
        \dot{s} &= \frac{v\cos{\tilde{\psi}}}{1-c(s)y},\\ 
		\dot{y} & = v\sin{\tilde{\psi}} + \dot{Y}_P,\quad\text{and} \\
    \dot{\tilde{\psi}}&= \dot{\psi}- c(s)\frac{v\cos{\tilde{\psi}}}{1-c(s)y},
	\end{aligned} 
 \label{eq:kinematicPointOWithSliding}
\end{equation}
with
\begin{equation}
    \begin{aligned}
        \dot{Y}_P &= v\cos{\tilde{\psi}}\tan\beta^R, \quad\text{and}\\
        \dot{\psi} &= v\frac{\tan\left(\delta+\beta^F\right) - \tan\beta^R}{L} \cos{\beta^R},
    \end{aligned}
    \label{eq:lateral_speed_disturbance}
\end{equation}
where $v$ denotes the velocity of the vehicle.
The variable $\dot{Y}_P$ is the lateral speed disturbance induced by the rear sideslip angle and $\dot{\psi}$ the rotational speed of the vehicle.
From the kinematics of the robot’s center, we infer in the next section the kinematics of the implement.

\subsubsection{Kinematic Model of the Offset Point $I$}
From the definitions in~\autoref{eq:RelationEcart}, the time derivative of the tracking error is expressed as
\begin{equation}
    \begin{aligned}
		\dot{e}_I & = \dot y + \dot{\tilde{\psi}} \left(I_s  \cos\tilde{\psi} - I_y \sin\tilde{\psi} + \dv{e}{\tilde{\psi}}\right),
	\end{aligned} 
 \label{eq:DotYT}
\end{equation}
where $\dot{e}_I$ denotes the time derivative, and
\begin{equation}
    \dv{e}{\tilde{\psi}} = -c(s)\frac{r_I\left(I_y \cos\tilde{\psi} - I_s \sin\tilde{\psi}\right)}{\sqrt{1-c(s)^2r_I^2}}, 
    \label{eq:dedt}
\end{equation}
with 
\begin{equation}
    \begin{aligned}
        r_I = I_s \cos\tilde{\psi} + I_y \sin\tilde{\psi}.
    \end{aligned}
\end{equation}
\autoref{eq:dedt} exists if $|I_s\cos\tilde{\psi} + I_y\sin\tilde{\psi}|<|\frac{1}{c(s)}|$, which is the same condition as~\autoref{eq:cond_TsTy}.

To simplify the expression of \autoref{eq:DotYT}, we make the assumption that the steering angle of agricultural vehicles is limited, and thus the curvature $c(s)$ of the reference path remains low. Therefore, we neglect terms whose order is equal to or greater than $c^2(s)$. 
Consequently, the term $\dot{\tilde\psi}\dv{e}{\tilde{\psi}}$ can be simplified to
\begin{equation}
    \dot{\tilde\psi}\dv{e}{\tilde{\psi}} \approx -c(s) \dot{\tilde\psi}r_I\left(I_y \cos\tilde{\psi} - I_s \sin\tilde{\psi}\right),
    \label{eq:dedt_first_approx}
\end{equation}

\noindent
and using \autoref{eq:kinematicPointOWithSliding}, the term $c(s) \dot{\tilde\psi}$ inside \autoref{eq:dedt_first_approx} becomes
\begin{equation}
    \begin{aligned}
        c(s)\dot{\tilde{\psi}}& = v\Bigg[c(s)\frac{\tan(\delta + \beta^F) - \tan \beta^R}{L} \cos\beta^R \\
        &  \qquad\qquad -\frac{c^2(s)\cos\tilde{\psi}}{1-c(s)y} \Bigg].
    \end{aligned} 
    \label{eq:dedt_second_approx}
\end{equation}
In~\autoref{eq:dedt_second_approx}, as the sideslip angles $\beta^R$ and $\beta^F$ are usually small~\cite{sideslipRoland}, the term $\tfrac{\tan(\delta + \beta^F) - \tan \beta^R}{L} \cos\beta^R$ is of the same order as $\tfrac{\tan\delta}{L}$, which represents the current curvature of the vehicle’s trajectory. Consequently, the product $c(s)\tfrac{\tan(\delta + \beta^F) - \tan \beta^R}{L} \cos\beta^R$ is of the same order as $c^2(s)$. 
\review{Typical agricultural tractors have a minimum turning radius of approximately $R_\text{min}=\SI{5.0}{\meter}$ \cite{Sarhan2010Steering}, which bounds the path curvature as $\abs{c(s)} \leq 1/R_\text{min}$ and, consequently, we have $c^2(s) \leq \SI{0.04}{\meter^{-2}}$. 
This implies a scaling factor of at least five between $c(s)$ and $c^2(s)$, which justifies neglecting higher-order terms in $c^2(s)$ and above.} 
Therefore,
\begin{equation}
    \begin{aligned}
        c(s)\dot{\tilde{\psi}}& \approx 0.
    \end{aligned} 
    \label{eq:dedt_third_approx}
\end{equation}
By substituting~\autoref{eq:dedt_third_approx} into~\autoref{eq:dedt_first_approx}, the term $\dot{\tilde\psi}\dv{e}{\tilde{\psi}}$ can be neglected compared to the other terms in~\autoref{eq:DotYT}, leading to
\begin{equation}
    \begin{aligned}
		\dot{e}_I & = \dot{y} + \dot{\tilde{\psi}} \left(I_s  \cos\tilde{\psi} - I_y \sin\tilde{\psi}\right).
	\end{aligned} 
 \label{eq:DotYTapprox}
\end{equation}
Finally, the overall kinematics of the system are given by 
\begin{equation}
    \begin{aligned}
        \dot{s} &= \frac{v\cos{\tilde{\psi}}}{1-c(s)y}, \\
                \dot{e}_I & = v\sin\tilde{\psi}+ \dot{Y}_P + \dot{\tilde{\psi}}\left(I_s\cos\tilde{\psi} - I_y\sin\tilde{\psi}\right),\\
        \dot{\tilde{\psi}}& = \dot{\psi} - c(s)\frac{v\cos\tilde{\psi}}{1-c(s)y},
    \end{aligned} 
    \label{eq:mainkinematic}
\end{equation}
both quantities remaining well-defined as long as $y\neq 1/c(s)$, indicating that a singularity arises when the position of the robot coincides with the center of the osculating circle. 
Nonetheless, given that the curvature does not reach high values in agricultural scenarios, owing to the limited maximum steering angle of the robot, the likelihood of encountering such a singularity is minimal as long as the robot is near the reference path. 

\subsubsection{Sliding Observer}
Sideslip angles $\beta^R$ and $\beta^F$ are not directly measurable, as there are no simple sensors allowing access to their values. 
\review{In this work, the sideslip angles are estimated using an observer-based approach, using the method described in \cite{sideslipRoland}. Note that this paper does not make any contribution to the observer design.}

The method estimates the sideslip angles such that the estimated lateral and angular deviations converge toward those measured by sensors such as~\ac{GNSS}.
It provides an estimation of sideslip angles $\hat{\beta}^R$ and $\hat{\beta}^F$ and, finally, the estimation of the lateral speed disturbance $\dot{\hat{Y}}_P$. 
These estimations are then considered to be representative of actual grip conditions and will then be used to compute the control law. 
\rereview{As in \cite{sideslipRoland}, the observer is designed to converge significantly faster than the control loop.}
In the following sections, the notations used for sideslip angles and lateral perturbations  ${\beta}^R$, ${\beta}^F$ and ${\dot{Y}}_P$ are considered equivalent to their corresponding estimated values.

\subsection{Control of an Offset Point}
In this section, we first present the direct adaptation of the control laws presented in \cite{offset_point_control_article} to account for the sliding observer.
Then, we build upon this work to create a predictive-based method to control an offset point on the robot.

\subsubsection{Lateral Servoing of the Robot's Center O}\label{directe_control_section}
Presented in detail in~\cite{offset_point_control_article}, this control law enables the control of the tracking error of point $I$ to zero (i.e. $e_I \rightarrow 0$) by regulating a non-null tracking error of the robot's center $O$, using classical path tracking control.
This method requires computing a desired tracking error $y^d$ to be used as a reference for the tracking error $y$, thereby ensuring the convergence of the implement tracking error $e_I$ to zero. 

\begin{figure}[htbt]
    \centering
    \includegraphics[width=\linewidth]{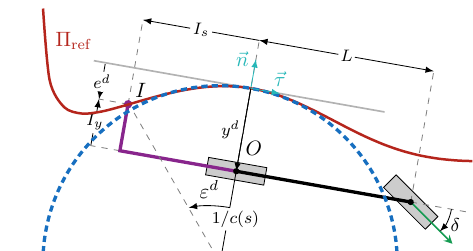}
    \caption{Representation of the offset of the point O when the point $I$ follows the reference path $\Pi_\text{ref}$. The desired tracking error of the point $O$ is set to $y_d$, enabling the implement $I$ to remain on the reference path. \review{Tracking errors are computed in the Frenet base $(\Vec{\tau}, \Vec{n})$.
    }
    }
    \label{fig:directe_control_scheme}
\end{figure}

As illustrated in~\autoref{fig:directe_control_scheme}, incorporating the offset into the desired tracking error leads the point $I$ to be aligned with the referenced path. 
This approach will be used as a baseline to evaluate our approach. The difference with the approach presented in~\cite{offset_point_control_article} is that sideslip angles will be considered in the classical control law of the point $O$ as presented in~\cite{Lenain2004AdaptiveAP}.

\subsubsection{Backstepping-based Control of the Point I}\label{backstepping_approach}
In this section, we present a backstepping control approach designed to directly regulate the implement tracking error $e_I$ based on the steering angle $\delta$. 
The developments presented in~\cite{offset_point_control_article} are revisited in this section, by using the model defined in~\autoref{eq:mainkinematic}, in which sideslip effects are accounted for.

The backstepping approach used here is divided into two stages: 
\begin{description}
    \item[S1] Determining the desired robot orientation $\tilde\psi^d$ to ensure the convergence of the tracking error of the implement $e_I$;
    \item[S2] Adjusting the steering angle of the robot $\delta$ to achieve the desired orientation $\tilde\psi^d$.
\end{description}
In the first stage, the objective is to obtain a desired angular deviation $\tilde{\psi}^d$ that ensures the convergence of the implement $I$ toward the reference path $\Pi_\text{ref}$. 
In this formulation, the angular rate $\dot{\tilde{\psi}}$ is not considered as actively controlled at this stage, but is assumed to be available through measurements.
It is important to highlight that, rather than relying on numerical differentiation of ${\tilde{\psi}}$, direct measurement using onboard sensors is preferred. 
To this end, the measured angular velocity is denoted as $\bar{\omega}$ throughout the remainder of this work for clarity.
Accordingly, the kinematic model governing the tracking error $e_I$ of the implement is reformulated in the initial stage of the backstepping design as follows:
\begin{equation}
    \dot{e}_I = v\sin\tilde{\psi}+\dot{Y}_P+\bar\omega\left(I_s\cos\tilde{\psi} - I_y\sin\tilde{\psi}\right).
    \label{eq:ytdot}
\end{equation}
The objective of the first backstepping stage is to determine a desired angular deviation $\tilde{\psi}^d$ that ensures exponential convergence of the tracking error $e_I$ to zero.

In~\autoref{eq:ytdot}, the derivative is taken regarding time.
Consequently, a control law derived from this equation would yield a convergence behavior characterized in terms of time.
In agricultural scenarios, a more relevant metric is the convergence distance:
in field operations, the main priority is to preserve crops, making theoretical guarantees on convergence distance particularly important.
Accordingly, by utilizing the expression of $\dot{s}$ and \autoref{eq:ytdot}, the kinematics of the tracking error of the implement $e_I$ can be reformulated with respect to the curvilinear abscissa as 
 \begin{equation}
\begin{aligned}
		e'_I &= \dv{e_I}{t}\left(\dv{s}{t}\right)^{-1} \\
             &= \alpha\left[\tan\tilde\psi+ \tan\beta^R+\gamma\left(I_s-I_y\tan\tilde\psi\right)\right],
\end{aligned} \label{eq:DerivYT}
\end{equation}
with
\begin{equation}
\begin{aligned}
    \alpha &= 1-c(s)y, \quad\text{and }\\
    \gamma &=\frac{\bar\omega}{v} \\
    &= \frac{\tan\left(\delta+\beta^F\right) - \tan\beta^R}{L} \cos{\beta^R} - \frac{c(s)\cos\tilde\psi}{\alpha}.
    \label{eq:defalphagamma}
    \end{aligned}
\end{equation}

\begin{figure}[htbp]
    \centering
    \includegraphics[width=\linewidth]{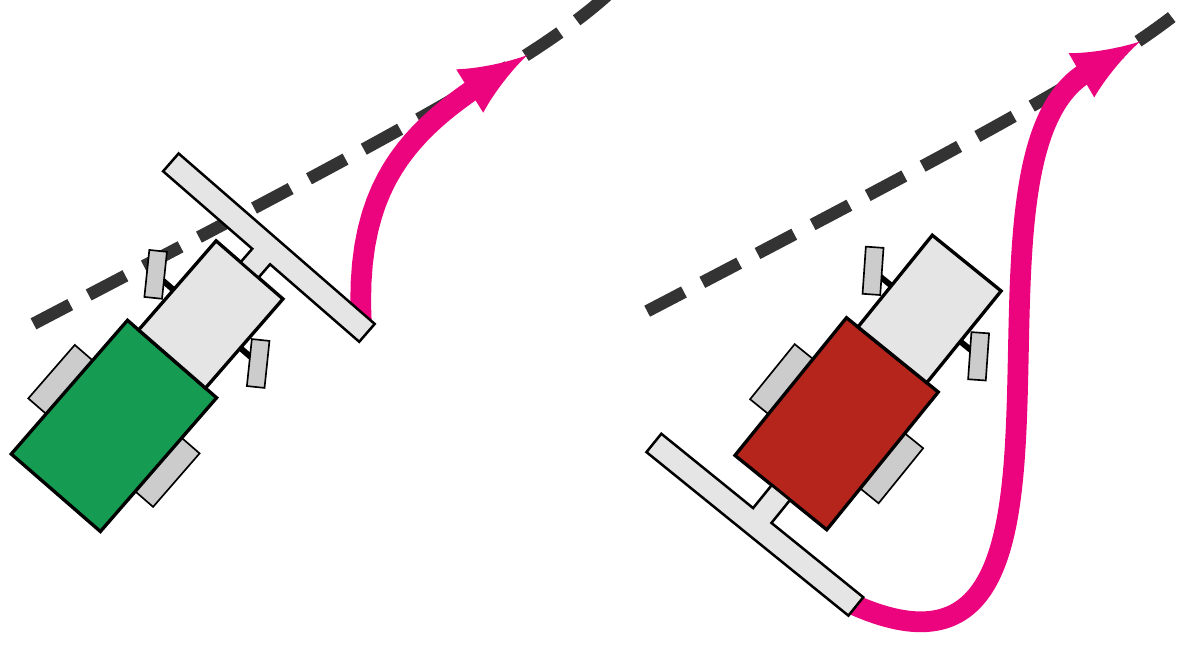}
    \caption{\review{Representation of the convergence process of an agricultural implement toward its reference path:
    Left: the implement is at the front, and is able to follow a pure exponential convergence.
    Right: the implement is located at the rear. To converge toward the reference path, it must first deviate from it, since the lever arm effect requires a vehicle rotation before convergence.}}
    \label{fig:forme_de_la_convergence_de_eI}
\end{figure}

In the case of an offset point at the rear of the robot, the kinematics of the point $I$ are such that when the vehicle steers to join the path, the point $I$ initially moves away from the reference path before gradually approaching it.
This phenomenon demonstrates that, to reduce the tracking error $e_I$ to zero, the convergence trajectory must follow a specific pattern, as illustrated in~\autoref{fig:forme_de_la_convergence_de_eI}. 

This behavior is typical of solutions to first-order differential equations with a second term that approaches zero at infinity 
\begin{equation}
\begin{aligned}
    y'(x) &= -ay(x) + f(x),
    \label{eq:equa_diff_de_convergence_example}
\end{aligned}
\end{equation}
with $a>0$  and $f(x)$ converges to zero.
Such a differential equation can fit both behaviors seen in \autoref{fig:forme_de_la_convergence_de_eI}.
Consequently, the convergence of the tracking error $e_I$ is ensured by formulating the following differential equation
\begin{equation}
\begin{aligned}
		e'_I & = -k_y e_I + \alpha \gamma I_s,
\end{aligned} \label{eq:Cond01}
\end{equation}
with $k_y>0$ a gain setting the convergence distance of the tracking error of the implement $e_I$. 
The convergence of the term $\alpha\gamma I_s$ to zero is guaranteed by the fact that $\gamma =\frac{\bar\omega}{v}$ approaches zero as the angular error and its derivative converge to zero.

Using the above constraint on $e_I'$ and \autoref{eq:DerivYT}, the desired angular deviation $\tilde\psi^d$ is expressed as
\begin{equation}
\begin{aligned}
		\tilde\psi^d & = \arctan \left(  \frac{-k_ye_I-\alpha\tan\beta^R}{\alpha(1-\gamma I_y)} \right).
\end{aligned} \label{eq:Step01}
\end{equation}
Two numerical singularities exist at $y=1/c(s)$, as also seen in~\autoref{eq:mainkinematic}, and at $1-\gamma I_y=0$. 
By rearranging terms, the latter singularity can be expressed as $v/\bar\omega = -I_y$.
This implies that when the control point lies at a distance equal to the turning radius of the vehicle, no orientation can satisfy~\autoref{eq:Cond01}.
However, given a relatively small velocity $v$ and an implement close to the vehicle, such a condition would require a very high rotational velocity, which once again is not realizable in the real-world conditions of agricultural machines.

If the angular deviation of the robot is equal to the value $\tilde\psi^d$, then the condition of~\autoref{eq:Cond01} is satisfied and the tracking error of the implement converges toward zero. 
Therefore, the remaining task is to ensure fast convergence of the angular deviation $\tilde{\psi}$ to its desired value $\tilde\psi^d$.

In the second backstepping stage, the error is defined as $e_\psi=\tilde\psi-\tilde\psi^d$. Neglecting the variations of $\tilde\psi^d$, the spatial derivative of this error can be expressed as
\begin{equation}
\begin{aligned}
		e'_\psi &= \dv{e_\psi}{t}\left(\dv{s}{t}\right)^{-1} \\
                  &\approx \dv{\tilde\psi}{t}\left(\dv{s}{t}\right)^{-1} \\
                  &= \frac{\alpha \left(\tan\left(\delta+\beta^F\right) - \tan\beta^R \right)}{L\cos{\tilde\psi}}\cos{\beta^R} - c(s).
\end{aligned} \label{eq:derivEtheta}
\end{equation}

\rereview{The variations of $\tilde\psi^d$ can be neglected because the desired steering dynamics are designed to converge faster than the desired angular deviation, allowing the second stage to treat the first stage input as quasi constant.}

The exponential convergence of the orientation of the vehicle toward its desired value is ensured by setting the following differential equation
\begin{equation}
\begin{aligned}
		e'_\psi & = -k_\psi e_\psi,
\end{aligned} \label{eq:Cond02}
\end{equation}
with $k_\psi>0$ the gain setting the convergence distance.

From \autoref{eq:derivEtheta} and the constraint given by \autoref{eq:Cond02}, the steering angle that ensures the convergence of the angular deviation to $\tilde\psi$ is given by
\begin{equation}
\begin{aligned}
		\delta^d & = \arctan \left(\frac{-k_\psi e_\psi + c(s)}{\alpha \cos{\beta^R}} L \cos{\tilde\psi} + \tan\beta^R\right) - \beta^F.
\end{aligned} \label{eq:ControlLawB}
\end{equation}
The angular deviation of the vehicle, will converge to its desired value $\tilde\psi^d$, and consequently, the implement point $I$ will exhibit exponential convergence toward the reference path $\Pi_\text{ref}$.

In summary, the implement tracking error $e_I$ can be computed from the tracking errors and the angular deviation $y,\tilde\psi$, as measured by onboard sensors, using~\autoref{eq:RelationEcart}. 
From this, the current desired angular deviation $\tilde\psi_d$ is determined using~\autoref{eq:Step01}, and the steering angle to be applied to the robot is then computed via~\autoref{eq:ControlLawB}, thereby guiding the implement toward the reference path.
Although this method enables the convergence of the point $I$ to the reference path, as seen in~\cite{offset_point_control_article}, there is no theoretical guarantee that it represents the optimal way to converge. 
Moreover, this approach is reactive and thus sensitive to local disturbances, particularly during curvature changes, as observed in~\cite{offset_point_control_article}.
If the reference path is a straight line, this method is sufficient. However, in the presence of curvature, changes in the path introduce disturbances that a purely reactive approach such as this cannot effectively handle due to actuator settling times.
To enhance the robustness of the method, a predictive control strategy is introduced in the first stage of the backstepping approach.

\subsubsection{\review {Predictive Control of the Offset Point}} \label{sec:method_pred}
In this section, we establish a novel predictive control framework for controlling an implement point $I$.
Let $s_R$ be the current curvilinear abscissa of the robot and $s_h$ the length of the prediction horizon. 
This horizon is sampled with $n_h+1$ points regularly spaced on $[0,s_h]$ with a step of $\Delta s=s_h/n_h$ such that the tracking error at the $k^{th}$ sampling point is given by $e_I^k$.

By neglecting the terms in $O(\Delta s^3)$, the tracking error for a position s+$\Delta s$ with $ \Delta s \in [0, s_h]$ is given by 
\begin{equation}
    \begin{aligned}
        e_I\left(s_R+\Delta s\right) = e_I(s_R)+ e_I'(s_R)\Delta s + \defensereview{\frac{1}{2}}e_I''(s_R) \Delta s^2.
    \end{aligned} \label{eq:eIfutur}
\end{equation}

The expression of $e_I'(s)$ is given by \autoref{eq:DerivYT}. 
By differentiating \autoref{eq:DotYTapprox} and neglecting the terms involving $\ddot{\tilde{\psi}}$ and $\tilde{\psi}^2$,  the spatial derivative $e''_I$ is then given by 
 \begin{equation}
\begin{aligned}
		e''_I(s) &= \dv[2]{e_I(s)}{t}\left(\dv{s}{t}\right)^{-2} \defensereview{- \dv{e_I(s)}{t} \left(\dv[2]{s}{t}\right)\left(\dv{s}{t}\right)^{-3} }  \\
                &\approx \dot{\tilde{\psi}} v \left( \cos{\tilde{\psi}} - \sin{\tilde{\psi}} \tan\beta^R\right) \left( \frac{\alpha^2}{v^2 \cos^2{\tilde{\psi}}} \right) \defensereview{- \dot{e}_I(s) \frac{\ddot{s}}{\dot{s}^{3}}} \\
             &= \frac{\alpha^2}{\cos\tilde{\psi}} \left(1-\tan\tilde{\psi}\tan\beta^R\right)\gamma \defensereview{- \dot{e}_I(s) \frac{\ddot{s}}{\dot{s}^{3}}},
\end{aligned} \label{eq:DerivDerivYT}
\end{equation}

with $\gamma$ given by~\autoref{eq:defalphagamma}\defensereview{, $\dot{e}_I(s)$ given by \autoref{eq:mainkinematic}, $\dot{s}$ given by \autoref{eq:kinematicPointOWithSliding} and 
\begin{equation}
\begin{aligned}
\ddot s
=
-\frac{v\sin(\tilde{\psi})\dot{\tilde{\psi}}}{1-cy}
+
\frac{cv\cos(\tilde{\psi})\dot y}{(1-cy)^2}
\end{aligned}
\end{equation}
where $\dot{\tilde{\psi}}$ and $\dot y$ are given by \autoref{eq:kinematicPointOWithSliding} }.
\defensereview{Note that the effect of the second term, $- \dot{e}_I(s) \frac{\ddot{s}}{\dot{s}^{3}}$ of the \autoref{eq:DerivDerivYT}, is found to be negligible during our experiments.}
The second derivative $e''_I$ is assumed to remain constant over the prediction horizon. 
Its value is computed using~\autoref{eq:DerivDerivYT} and is taken as the one corresponding to the end of the prediction horizon. 
This choice leads to better anticipating curvature changes.
As such, $e_I^k$ is then given by 
\begin{equation}
    \begin{aligned}
        e_I^k = e_I (s_R)+ e_I'(s_R) k\Delta s + \defensereview{\frac{1}{2}} e''_I(s_R)k^2\Delta s^2.
    \end{aligned} \label{eq:PredictiveControlLaw2}
\end{equation}

\begin{figure}[htbp]
\centering
\includegraphics[width=\linewidth]{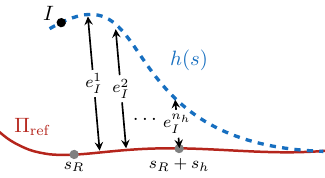}
\caption{\rereview{Illustration of the proposed approach:
the predictive control input is defined as the one that minimizes the tracking error between the reference path $\Pi_\text{ref}$ and the chosen convergence trajectory profile $h(s)$ over a given prediction horizon $s_h$.} }
\label{fig:optimal_principe}
\end{figure}

The predictive control scheme determines the angular deviation $\tilde{\psi}$ that minimizes the cumulative tracking errors $e_I^k,\,0 < k \leq n_h$ over the prediction horizon $s_h$, as illustrated in \autoref{fig:optimal_principe}.

The optimization criterion $J$ is defined as
\begin{equation}
    \begin{aligned}
        J &= \sum_{k=1}^{n_h}\left[e_I^k - h(k \Delta s)) \right]^2,
    \end{aligned} \label{eq:optimisatiocriterion}
\end{equation}
with 
\begin{equation}
    h(k\Delta s) = e_I(s_R)e^{-\lambda k \Delta s} + \alpha \gamma I_s k \Delta s.
\end{equation}
The form of equation that can satisfy both cases depicted in \autoref{fig:forme_de_la_convergence_de_eI}.
\rereview{Note that it is a design choice to ensure convergence in both cases shown in \autoref{fig:forme_de_la_convergence_de_eI}.}
This convergence rate can be adjusted using the parameter $\lambda >0$.
The criterion $J$ thus represents the sum of the deviations between the tracking error of the implement and the reference convergence function. 
Developing \autoref{eq:optimisatiocriterion}, we obtain
\begin{equation}
    \begin{aligned}
        J = \sum_{k=1}^{n_h}\big[e_I(s_R) + \left(\xi+ A\right)k\Delta s &+ \defensereview{\frac{1}{2}}  e''_I(s_R)k^2\Delta s^2 \\ 
        &- e_I(s_R)e^{-\lambda k \Delta s})\big]^2,
    \end{aligned} 
    \label{eq:optimisatiocriterion1}
\end{equation}
with the following change of variable,
\begin{equation}
    \begin{aligned}
         \xi &= \alpha \left(1-\gamma I_y\right)\tan\tilde\psi, \quad\text{and}\\
         A &= \alpha \tan \beta^R. 
    \end{aligned} \label{eq:changementvariable}
\end{equation}
%
\review{Setting $\frac{\partial J}{\partial \xi} = 0$, the optimal value $\xi_d^h$ minimizing \autoref{eq:optimisatiocriterion1} is}
\begin{equation}
    \begin{aligned}
            \xi_d^h = - \frac{1}{\sigma_2} \left[e_I(s_R)\sigma_1 + A\sigma_2 +   e''_I(s_R) \sigma_3-e_I(s_R)\sigma_e\right], \\
    \end{aligned} 
    \label{eq:PredictiveControlLaw3}
\end{equation}
and thus 
\begin{equation}
    \begin{aligned}
            \tilde\psi_d^h = \arctan\left(\frac{\xi_d^h}{\alpha\left(1-\gamma I_y\right)}\right),
    \end{aligned} 
    \label{eq:optimal_psi}
\end{equation}
with
\begin{equation}
    \begin{aligned}
        \begin{array}{llll}
            \sigma_1 = \sum_{k=1}^{n_h} (k\Delta s)  \\
            \sigma_2 = \sum_{k=1}^{n_h} (k\Delta s)^2  \\
            \sigma_3 = \defensereview{\frac{1}{2}} \sum_{k=1}^{n_h} (k\Delta s)^3 \\
            \sigma_e = \sum_{k=1}^{n_h} \left(k\Delta se^{-\lambda k\Delta s}\right).\\
        \end{array}
    \end{aligned} \label{eq:PredictiveControlLaw3}
\end{equation}
This desired orientation exists if $\alpha \neq 0$ and $1-\gamma I_y \neq 0$. 
These conditions are the same as those in~\autoref{eq:Step01}.
Furthermore, one can notice that for $n_h = 1$ and $\Delta s$ small (\ie no prediction), the optimum value $\xi_d^h$ is 
\begin{equation}
    \begin{aligned}
        \xi_d^h &=  -e_I(s_R) \frac{1-e^{-\lambda \Delta s}}{\Delta s} - \alpha \tan \beta ^R \\
                &\approx - \lambda e_I(s_R) - \alpha \tan \beta ^R, 
    \end{aligned}
\end{equation}
 \review{and thus the optimal angular deviation minimizing \autoref{eq:optimisatiocriterion1} is}
 \begin{equation}
    \begin{aligned}
        \tilde\psi_d^h = \arctan\left(\frac{-\lambda e_I(s_R) - \alpha\tan \beta^R }{\alpha(1-\gamma I_y)}\right),
    \end{aligned} 
 \end{equation}
being exactly the backstepping expression of the desired angular deviation obtained in \autoref{eq:Step01}, providing the choice for the desired convergence distance, tuned by parameter $\lambda$ is equivalent to the choice for the parameter $k_y$.
As such, in the absence of prediction, the predictive method is then equivalent to the backstepping approach.

Once the optimal angular deviation $\tilde\psi_d^h$ is found via \autoref{eq:optimal_psi}, the resulting steering angle is computed the same way as done in the backstepping approach using \autoref{eq:ControlLawB}.
\review{This approach combines the advantages of backstepping control presented in \cite{offset_point_control_article}, such as the regulation of intermediate variables like the orientation of the vehicle, with a predictive component that anticipates curvature changes along the reference path.}

\section{SIMULATIONS}  \label{sec:simulation}
\review{Before conducting real-world experiments, this section focuses on a simulation-based validation of our novel predictive control strategy. 
First, we evaluate the impact of the robot's velocity on the path tracking error, for an implement at the front and rear of the robot.
Second, we investigate the influence of the offset point position on the tracking error.
An Ackermann-steering vehicle is modeled, and lateral tire slip effects are simulated in the system dynamics using the linear Pacejka tire model \cite{article_Park}. 
The cornering stiffness coefficient was set to $C=\SI{7 500}{\newton\per\radian}$, which is representative of wet agricultural soil and consistent with the cornering stiffness values used in \cite{article_Han}.}
\review{In the following, a set of 17 representative trajectories with varying curvatures, lengths, and shapes similar to those described in \cite{agriengineering4040075} is used for the evaluation. 
}

\begin{figure}[ht]
    \centering
    \includegraphics[width=\linewidth]{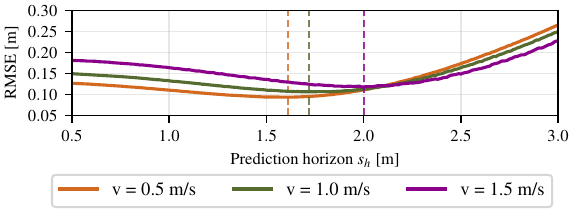}
    \caption{\rereview{Evolution of the \ac{RMSE} of the implement tracking error with respect to the prediction horizon $s_h$ for different velocities with their associated optimal values (vertical dashed lines).}}
    \label{fig:square_error_vs_sh}
\end{figure}

\review{
The steering actuator response time is set to $\SI{1}{\second}$ to match the subsequent real experiments, which limits the steering-angle convergence to no less than approximately $\SI{2}{\second}$. 
At a vehicle speed of $v = \SI{2}{\meter\per\second}$, this corresponds to a minimum convergence distance of approximately $\SI{4}{\meter}$. Accordingly, a conservative convergence distance of $\SI{5}{\meter}$ is selected for the angular deviation, yielding a gain of $k_\psi = 0.6$.
Since the convergence of the tracking error $e_I$ is governed by a slower outer loop, its convergence distance is chosen four times larger, resulting in a distance of $\SI{20}{\meter}$. This corresponds to a gain value of $\lambda = 0.15$.
For each implement position and velocity, the prediction horizon $s_h$ is determined through an
optimization process performed on a separate set of trajectories that are not used
for the evaluation phase.
} 
\rereview{
The optimization objective function is defined as the \ac{RMSE} of the implement tracking error, and the optimization is carried out using a grid search algorithm. The prediction horizon $s_h$ is discretized within the interval $[0.5,\,3]\,\SI{}{\m}$ using a step size of $\SI{1}{\centi\meter}$.
}
\rereview{
As shown in~\autoref{fig:square_error_vs_sh}, for speeds of $\SI{0.5}{\meter\per\second}$, $\SI{1.0}{\meter\per\second}$, and $\SI{1.5}{\meter\per\second}$, an optimal value of $s_h$ can be identified. When $s_h$ is too small, the preview horizon remains limited and the controller relies on very local information, which reduces its predictive capacity. When $s_h$ is too large, the constant curvature assumption used in the prediction becomes less accurate, leading to degraded tracking performance.
Around the optimal value, the error curve exhibits a relatively flat region, indicating that the performance remains close to optimal at \SI{5}{\percent} over an interval of \SI{0.5}{\m}, for all depicted velocities. This suggests that no extensive tuning is necessary during deployments.
}

\subsection{Influence of the velocity}
    \review{
    First, we analyze the impact of the velocity on the tracking error. 
    Four representative offset point positions are evaluated, combining front or rear placement ($I_s = \pm \SI{2.0}{\meter}$) with left or right placement ($I_y = \pm \SI{0.5}{\meter}$).
    Eight different velocities are tested, ranging from $\SI{0.25}{\meter\per\second}$ to $\SI{2.0}{\meter\per\second}$ with an increment of $\SI{0.25}{\meter\per\second}$, representative of agricultural operations \cite{agriculture_Wen}. This results in a total of $17\!\times\! 4\! \times\! 8\!  = 544$ simulation runs for this experiment.
    }
 
    \begin{figure}[h]
        \centering
        \includegraphics[width=\linewidth]{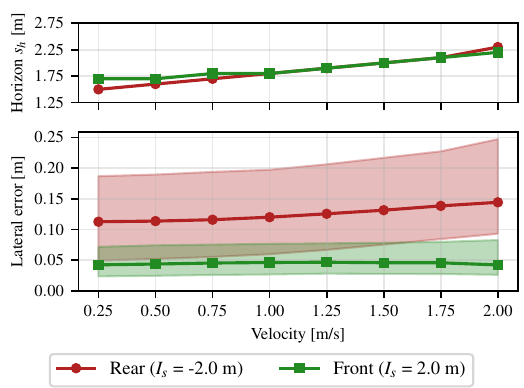}
        \caption{\review{Bottom: Evolution of the median tracking error and interquartile range of the implement point as a function of speed. 
        Each point corresponds to the median tracking error values throughout all the simulated trajectories. 
        The shaded area represents the interquartile range at the \SI{25}{\%} and \SI{75}{\%} percentiles.
        Top: optimal prediction horizon $s_h$ used for each speed.}}
        \label{fig:erreur_en_fonction_vitesse}
    \end{figure}
    \review{The results are presented in \autoref{fig:erreur_en_fonction_vitesse}, depicting the tracking error as well as the optimal horizon of prediction for each velocity. 
    By mirror symmetry with respect to the vehicle longitudinal axis, a right turn with a left-mounted implement is equivalent to a left turn with a right-mounted implement.
    Consequently, the Front-Left and Front-Right configurations are grouped together, as are the Rear-Left and Rear-Right configurations.
    The median and interquartile range of the tracking error of the implement point, computed over the full set of reference paths, are presented as functions of vehicle speed.
    Overall, the tracking error of the implement point remains within acceptable limits regardless of vehicle speed, trajectory type, or implement offset, with all median values remaining below \SI{15}{\centi\meter}. However, both the median and the interquartile range of the tracking error are larger when the implement is mounted at the rear of the vehicle compared to a front-mounted configuration. This behavior can be
    attributed to the combined effect of the longer lever arm and the infeasibility of certain reference paths involving curvature discontinuities, which lead to increased tracking errors during transient phases.
    Furthermore, the optimal prediction horizon $s_h$ increases with vehicle speed:
    as the forward speed of the vehicle increases, a longer prediction horizon becomes necessary.
    The median and interquartile range of the tracking error remain nearly constant as long as the prediction horizon $s_h$ is properly optimized. This behavior is particularly evident for front-mounted implements, while rear-mounted configurations have a slight increase in tracking error with forward speed.
    }

\subsection{Influence of the Offset Point Position}
    \review{In this section, we investigate the influence of the implement position on the tracking error.
    The velocity of the robot is set to $\SI{1.0}{\meter\per\second}$, consistent with speeds commonly adopted for autonomous agricultural robots \cite{agriculture_Wen}.
    The implement position is sampled on both axes within the range $\SI{-3.0}{\meter}$ to $\SI{3.0}{\meter}$, with an increment of $\SI{0.25}{\meter}$, resulting in $17\!\times\! 25^2\!   = 10625$ simulation runs.
    This range is representative of the typical geometric offsets encountered for agricultural implements on autonomous robots.
    }

    \begin{figure*}[b]
        \centering
        \includegraphics[width=\linewidth]{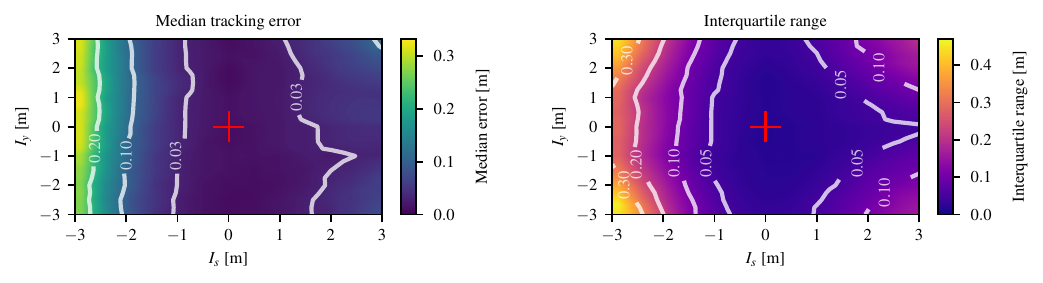}
        \caption{\review{Left: Median tracking error as a function of its longitudinal and lateral offsets of the implement control point. Right: Associated interquartile range at \SI{25}{\%} and at \SI{75}{\%}.
        $I_s$ denotes front ($>0$) or rear ($<0$) placement, $I_y$ left ($>0$) or right ($<0$); the origin $(0,0)$ marks the midpoint of the rear axle (red cross) and corresponds to a classical control framework.
        }
        }  
        \label{fig:error_colormap_ts_ty}
    \end{figure*}

    \review{The results are presented in \autoref{fig:error_colormap_ts_ty}:
    at the origin ($I_s\!=\!I_y\!=\!0$), the problem reduces to the classical case of controlling the center of a mobile robot, and the median tracking error is as expected, minimal, as well as the interquartile range. 
    In this configuration, the tracking error of the implement point indeed remains close to zero, with a median value of approximately \SI{2}{\centi\meter}, which is consistent with the results reported in \cite{sideslipRoland}.
    A symmetry with respect to the axis $I_y = 0$ can be observed in the evolution of the median tracking error. This indicates that the control performance is insensitive to whether the implement control point is located on the left or the right side of the vehicle.
    As $I_s$ moves away from the origin (front or rear), a progressive increase in the median tracking error is observed. 
    This increase is more pronounced when the implement is at the rear ($I_s < 0$).
    This behavior is attributed to the lever-arm effect, which becomes more pronounced as the implement moves away from the robot's center, thereby amplifying the tracking error during the transient phases.
    %
    The interquartile range reveals a stronger degradation for rear offsets, coming from the tracking erros during curvature changes. 
    This phenomenon can be explained by the behavior illustrated in \autoref{fig:forme_de_la_convergence_de_eI}: when the offset point is positioned behind the vehicle's center, it initially deviates from the reference path before converging, leading to increased tracking error during the transient phases.
    }

\section{EXPERIMENTS}\label{sec:exp}
In this section, we present real-world experiments under various conditions to validate our approach. 
First, the experimental setup is described, followed by real-world experiments to compare and evaluate the proposed method against state-of-the-art methods \cite{offset_point_control_article} and \cite{lukassek_model_2020}. 
Finally, we provide an analysis of our method in a real-world agricultural scenario.
\review{The control-law validation experiments were conducted on grass, with the implement raised and not in contact with the grass, whereas the agricultural scenario was performed on cultivated soil with the implement in contact with the soil. This setup validates the proposed framework under standard agricultural operating conditions.}
\begin{figure}[htbp]
    \centering
    \includegraphics[width=\linewidth]{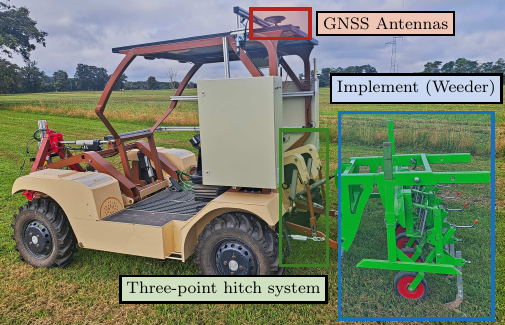}
    \caption{Robotic platform used in this paper. It is equipped with two GNSS antennas to accurately estimate the position and orientation of the vehicle.
    A weeder is rigidly attached at the rear of the robot.}
    \label{fig:plateforme_robotique}
\end{figure}

\subsection{Experimental Setup}
The following experiments were conducted using the robot depicted in the~\autoref{fig:plateforme_robotique}. The platform is a four-wheel-drive, Ackermann-steering vehicle capable of towing various implements, such as mechanical weeders.
It is equipped with two~\ac{RTK}~\ac{GNSS} sensors coupled with an~\ac{IMU}, which enables the measurement of the state vector as well as angular and longitudinal velocities. 
\review{The implementation was carried out in C/C++ using the \ac{ROS} middleware. The embedded computing platform is based on an \href{https://www.neousys-tech.com/en/product/product-lines/edge-ai-gpu-computing/nuvo-9160gc-intel-12th-gpu-computing-platform}{Intel\textsuperscript{\textregistered} Core\textsuperscript{\texttrademark} i5-9500E} processor, featuring 12 threads and a base frequency of 2.9\,GHz.
}

\review{Regarding the tuning of the control law parameters, the values obtained from simulation were used as an initial baseline and subsequently fine-tuned manually on short trajectories.
During the tuning process, it was observed that a short prediction horizon $s_h$ leads to a more reactive control action, whereas an excessively long horizon causes the controller to over-anticipate, thereby increasing the tracking error.
However, note that the tracking error is only weakly affected by the prediction horizon, and as long as an appropriate value is used no fine-tuning is necessary.
}
\begin{figure*}[b]
    \centering
    \includegraphics[width=\linewidth]{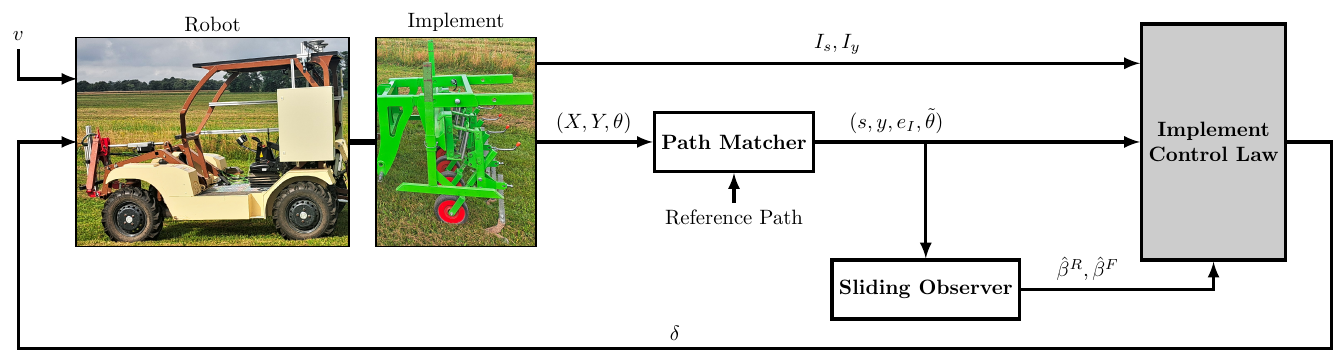}
    \caption{Pipeline of the system. 
    Using onboard sensors, the positions of both the robot and the implement are known. A path-matching algorithm, based on the reference path and the current position of the robot, computes the tracking errors. These errors are then used to estimate the sideslip angles, which are subsequently incorporated into the control law to compute the required steering angle for the robot.
    }
    \label{fig:diagram_bloc}
\end{figure*}
An overview of the system architecture is provided in~\autoref{fig:diagram_bloc}.
Using the \ac{GNSS} receivers and the \ac{IMU}, the position of the center of the rear axle and the orientation of the robot are estimated in real time. 
The position of the implement is then entirely known in an absolute frame. 
Using path-matching algorithms, the tracking error $y$, and the angular deviation  $\tilde{\psi}$ of the robot and the implement tracking error $e_I$ are computed, regarding a preliminarily defined path. 
These errors are then fed into the observer described in~\cite{sideslipRoland}, which estimates the sideslip angles $\hat\beta^R$ and $\hat\beta^F$. 
The estimated sideslip angles are subsequently used by the controller to compute the control inputs required to ensure that the implement accurately follows the reference path.
The robot operates at a constant velocity of \SI{1.0}{\m\per\s} for all experiments.
\review{The observer reduces the tracking error of the implement by approximately \SI{50}{\%} to \SI{64}{\%}, particularly along curved paths. 
This improvement can be attributed to increased wheel slip caused by natural ground conditions (\eg grass, agricultural soil), which was aggravated by light rainfall before the experiments.}

\subsection{Validation of the Control Law}
\begin{figure}[htbp] 
      \centering
       \includegraphics[width=\linewidth]{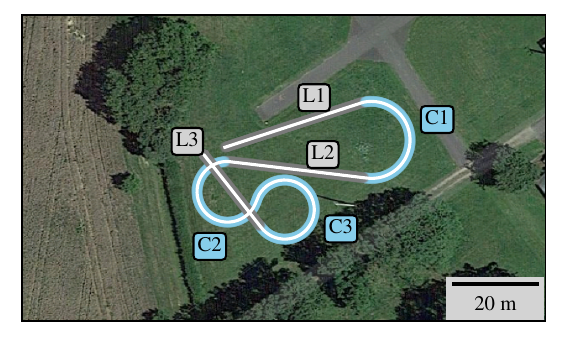}
      \caption{Reference path used in the first part of the experiments. 
      It consists of three straight segments $(L1, L2, L3)$ and three circular arcs $(C1, C2, C3)$ with different curvatures. It starts with the straight line $L1$ and ends with $L3$.
      This path features all transition types, which are straight-to-arc, arc-to-straight, and arc-to-arc.}
      \label{fig:trajries}
\end{figure}
In this section, we investigate the effectiveness of our method and compare its performance against state-of-the-art approaches.
The reference path is depicted in~\autoref{fig:trajries}. 
The experiments were conducted shortly after a light rainfall, which left the grass slightly damp, introducing wheel slippage.
The reference path is composed of three straight lines and three circular arcs. 
The arrangement of these segments enables the observation of standard curvature transitions, including straight-to-arc, arc-to-straight, and arc-to-arc.

\subsubsection{Impact of the implement's position}
\review{
The performance of the proposed predictive control law is evaluated on the reference path in~\autoref{fig:trajries}, which combines straight and circular segments with line-to-arc, arc-to-arc, and arc-to-line transitions, under various implement configurations.%
}
Specifically, we consider four spatial arrangements of the implement relative to the robot: front-right, front-left, rear-right, and rear-left. 
These positions correspond to the standard size of agricultural implements.
The controller parameters associated with each configuration are summarized in~\autoref{tab:parameters_expe2}.  
\newcommand{\coloredbar}[2]{%
  \textcolor{#1}{\rule[-.9em]{2.5pt}{2.5em}} 
  \hspace{0em}#2%
}
\definecolor{color_fr}{HTML}{4B0082}
\definecolor{color_fl}{HTML}{D2691E}
\definecolor{color_rr}{HTML}{008B8B}
\definecolor{color_rl}{HTML}{6B8E23}
\begin{table}[htbp]
    \caption{Parameters used in the validation of the control law experiment}
    \label{tab:parameters_expe2}
    \centering
    \begin{tabular}{l l l}
        \toprule
       &  \textbf{Implement
       position}   & \textbf{Parameters}\\
        \midrule
       \multirow{2}{*}{\coloredbar{color_fl}{Front-Left (FL)}}     &    $I_s = \SI{2.0}{\m}$ & $\lambda = 0.15$, $k_\theta = 0.4 $ \\
              &   $I_y = \SI{0.5}{\m}$                  & $s_h = \SI{0.5}{\m}$\\
       \multirow{2}{*}{\coloredbar{color_fr}{Front-Right (FR)}}       &    $I_s = \SI{2.0}{\m}$ & $\lambda = 0.15$, $k_\theta = 0.4 $ \\
              &   $I_y = \SI{-0.5}{\m}$                  & $s_h = \SI{0.5}{\m}$ \\
       \multirow{2}{*}{\coloredbar{color_rl}{Rear-Left (RL)}}      &    $I_s = \SI{-2.0}{\m}$ & $\lambda = 0.2$, $k_\theta = 0.8 $ \\
              &   $I_y = \SI{0.5}{\m}$                  & $s_h = \SI{2.0}{\m}$\\
        \multirow{2}{*}{\coloredbar{color_rr}{Rear-Right (RR)}}       &    $I_s = \SI{-2.0}{\m}$ & $\lambda = 0.2$, $k_\theta = 0.8 $ \\
              &   $I_y = \SI{-0.5}{\m}$                  & $s_h = \SI{2.0}{\m}$\\
        \bottomrule
    \end{tabular}
\end{table}
\begin{figure*}[thbp]
    \centering
    \includegraphics[width=\linewidth]{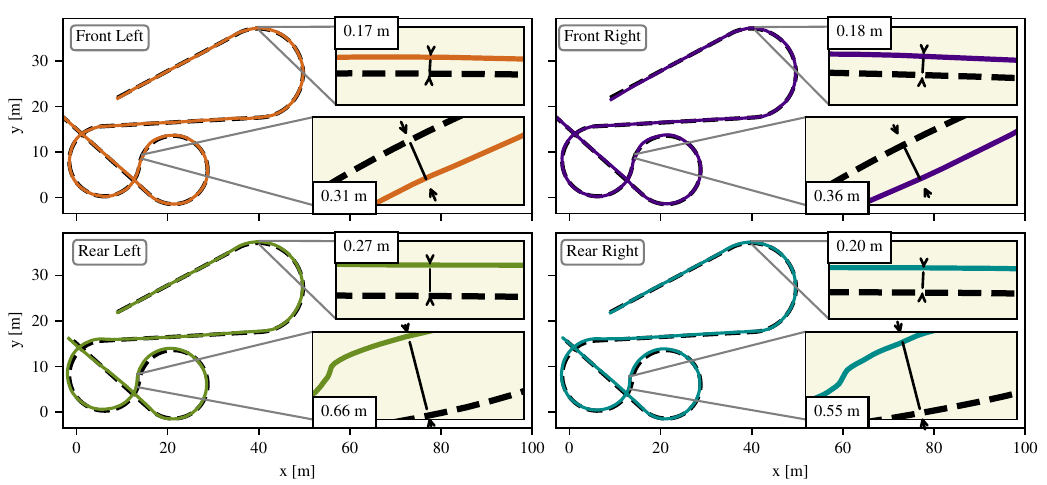}
    \caption{Control Law Validation with Various Implement Positions: Front-Left, and Front-Right, Rear-Left and Rear-Right.
    Each plot shows both the reference path in dashed black and the trajectory followed by the implement. 
    Two zoom-in views are presented:
    The first highlights the peak tracking error of the implement point during the first transition phase, which corresponds to a straight-line to circular-arc segment. The second one highlights the peak tracking error of the implement during the fourth transition phase, which corresponds to a circular-arc to circular-arc segment.
    Note that, because of the lever arm effect of the implement, the reference path is physically infeasible in some parts and the tracking error cannot be reduced to zero.
    }
      \label{fig:control_law_validation}
\end{figure*}

As shown in~\autoref{fig:control_law_validation}, the implement converges toward the reference path regardless of its position (front-left, front-right, rear-left, or rear-right). However, due to the infeasibility of the trajectory, tracking errors inevitably appear during curvature changes, especially when curvature is high and control point is situated at the rear of the robot.
\autoref{fig:control_law_validation} highlights two transition phases (line-to-arc and arc-to-arc). It can be observed that, when the implement is mounted at the front, the tracking error during the line-to-arc transition reaches $\SI{17}{\centi\meter}$ for the front-left position and $\SI{18}{\centi\meter}$ for the front-right position. During the more challenging arc-to-arc transition, the maximal tracking error increases to $\SI{31}{\centi\meter}$ for the front-left position and $\SI{36}{\centi\meter}$ for the front-right position.
Conversely, when the implement is mounted at the rear, the tracking error during the line-to-arc transition is up to $\SI{27}{\centi\meter}$ for the rear-left position and $\SI{20}{\centi\meter}$ for the rear-right position. For the most demanding arc-to-arc transition, tracking errors of $\SI{66}{\centi\meter}$ and $\SI{55}{\centi\meter}$ are observed for the rear-left and rear-right positions, respectively.

When the implement is mounted at the rear, larger transient tracking errors occur due to a combination of the lever-arm effect and the infeasibility of the path, which contains abrupt curvature changes that the implement cannot physically follow. In contrast, when the implement is mounted at the front, its position relative to the robot’s rear axle provides an effective anticipation of the path, naturally mitigating tracking errors.

\subsubsection{Approaches Comparison}
The previous experiments demonstrate that the proposed approach enables accurate path tracking of the implement regardless of its position, although some tracking errors are inevitable during transition phases. Importantly, these tracking errors cannot be entirely eliminated, as the reference trajectories are by nature infeasible: the same path must be followed with implements of different sizes and lever arms, making it impossible to design a universally feasible trajectory. The strength of the proposed approach lies in its ability to minimize these unavoidable tracking errors, outperforming purely reactive strategies. To highlight this fact, we extend the analysis by comparing our method with state-of-the-art approaches. To the best of our knowledge, the leading state-of-the-art methods addressing the control of an offset point are 
\begin{itemize}
    \item A Lateral offset Servoing Control~\cite{offset_point_control_article}, with sideslip angles taken into account.
    \review{ In this method, the reference path is tracked by the implement point through the assignment of a desired tracking error at the robot’s center.}
    \item A Backstepping Control, introduced in~\cite{offset_point_control_article}, accounting for sideslip effects as described in the previous section;  and
    \item A \acf{MPC} method designed for front-mounted implements~\cite{lukassek_model_2020}. 
    \review{ This approach consists of computing the control input that minimizes the distance between the implement point and its corresponding matched point over the prediction horizon; however, it is formulated under the assumption that the implement is located at the front of the vehicle.}
\end{itemize}
\review{The Lateral Offset Servoing Control and the Backstepping Control proposed in \cite{offset_point_control_article} are both reactive approaches. 
The Lateral Offset Servoing Control is based on an existing control law applied to the robot center, combined with a geometric analysis that allows a desired tracking error to be assigned to the robot center. However, this construction does not provide theoretical convergence guarantees comparable to those offered by the backstepping approach.
The \acf{MPC} method proposed in \cite{lukassek_model_2020} is predictive in nature, but it is specifically designed for scenarios involving front-mounted implements, in contrast to the two aforementioned reactive methods.
 } 

In the following experiment, and to remain consistent with the configurations used in the state-of-the-art method presented in~\cite{lukassek_model_2020}, the implement is positioned at the front ($I_s = \SI{2.0}{\m}, I_y = \SI{0.5}{\m}$).
The reference path used is the one depicted in~\autoref{fig:trajries}.
The parameters used for each method are summarized in~\autoref{tab:parameters_expe3}.
\begin{table}[thbp]
    \caption{Parameters used in the comparison experiment. All parameters have been hand-tuned to reach the best possible results on the given trajectory.}
    \label{tab:parameters_expe3}
    \centering
    \begin{tabularx}{\linewidth}{p{2.5cm}X}
        \toprule
    \textbf{Method}   & \textbf{Parameters}\\
        \midrule
    Lateral Servoing~\cite{offset_point_control_article} & $k_d = 0.7$, $k_p = 0.13$ \\
    Backstepping~\cite{offset_point_control_article}     & $k_y = 0.15$, $k_\theta = 0.6$ \\
    \ac{MPC}~\cite{lukassek_model_2020}           & $Q = [1, 5, 1]$, $R = [0.15 , 1]$, $h = \SI{1.0}{\s}$ \\
    \review{Predictive} (Ours)           & $\lambda = 0.15$, $k_\theta = 0.4 $ , $s_h = \SI{0.5}{\m}$\\
        \bottomrule
    \end{tabularx}
\end{table}

The method proposed by~\cite{lukassek_model_2020} is specifically designed for scenarios where the implement is mounted at the front of the vehicle. When the implement is positioned at the rear, the method fails to converge. This behavior can be explained by the fact that, in their~\ac{MPC} formulation, the objective function is constructed solely based on the tracking error of the implement, without enforcing a specific convergence profile.  
As shown in \autoref{fig:forme_de_la_convergence_de_eI} for the rear-mounted case, when the implement initially begins to diverge, the~\ac{MPC} approach proposed in~\cite{lukassek_model_2020} fails to anticipate that the tracking error must initially increase before it can converge. As a result, the controller steers sharply in the opposite direction in an attempt to immediately bring the implement back onto the reference path. This behavior increases the angular deviation and ultimately causes the system to diverge. To ensure a consistent comparison, we therefore restrict the analysis to the front-mounted implement case.
\begin{figure*}[htbp]
      \centering
       \includegraphics[width=\linewidth]{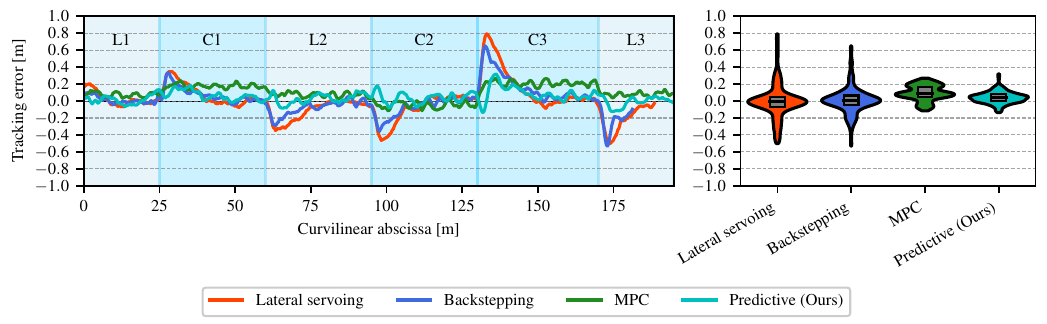}
      \caption{Comparison of the proposed approach with the three state-of-the-art methods in \cite{offset_point_control_article, lukassek_model_2020}.
      The straight segments (L1, L2, L3) and circular sections (C1, C2, C3) correspond to the areas highlighted in \autoref{fig:trajries}. 
      In the graph, the right-hand plot shows the overall distribution of tracking errors for each method, with the median and the $1^{st}$ and $3^{rd}$ quartiles represented as box plots.
      }
      \label{fig:approaches_comparison}
\end{figure*}

As depicted in~\autoref{fig:approaches_comparison}, all methods demonstrate convergence of the implement’s tracking error towards zero. 
However, significant differences arise during the transient phases. 
In these phases, our predictive control method outperforms the others by effectively minimizing implement tracking error. 
This effect is particularly noticeable during the transition between sections $C2$ and $C3$. 
Both sections are circular but have opposite curvatures, resulting in a sharp transition that is poorly handled by the reactive method presented in~\cite{offset_point_control_article}, leading to a significant offset. 
In contrast, this transition is better managed by predictive approaches, including the one proposed in this paper and the one developed in~\cite{lukassek_model_2020}. 
It is worth noting that the method proposed in~\cite{lukassek_model_2020} exhibits a higher and nearly constant offset during curved path segments ($C1$, $C2$ and $C3$). This behavior can be attributed to the fact that bad grip conditions (i.e. sideslip angles) are neglected in their proposed method.

A quantitative comparison is provided in~\autoref{tab:quantitative_comparison_expe3}, using the following performance metrics: 
\begin{itemize}
    \item The median tracking error of the implement, along with the \ac{IQR}. 
    These metrics characterize the overall performance of each method over the full trajectory.
    
    \item The maximum tracking error observed during each transient phase (transition between the segments depicted in \autoref{fig:trajries}), assessing how each method handles transitions between sections. 
    We examine the maximum tracking error within a window of $\SI{10}{\meter}$ before and after each transition point.
\end{itemize}
\begin{table}[thbp]
    \caption{Quantitative comparison of each method, including the median and interquartile range of the implement tracking error, as well as the maximum tracking error observed in each transient phase.
    The unit in meters ($\SI{}{\m}$). ``t.p.'' stands for transient phase.}
    \label{tab:quantitative_comparison_expe3}
    \centering
    \begin{tabularx}{\linewidth}{p{2cm}XXXX}
        \toprule
    \textbf{Implement tracking error}   & \textbf{Lateral Servoing} \cite{offset_point_control_article}& \textbf{Back stepping} \cite{offset_point_control_article}&\textbf{MPC} \cite{lukassek_model_2020}& \textbf{\review{Predictive} (Ours)}\\
        \midrule
    \textbf{Median} &  0.06 & 0.06 & 0.09 & \textbf{0.04} \\
    \textbf{IQR} &  0.18 & 0.11 & 0.11 & \textbf{0.06} \\
    \midrule
    \textbf{$1^{st}$ t.p. max} &  0.34 & 0.33 & 0.24 & \textbf{0.17} \\
    \textbf{$2^{nd}$ t.p. max} &  0.35 & 0.29 & 0.23 & \textbf{0.12} \\
    \textbf{$3^{rd}$ t.p. max} &  0.46 & 0.34 & \textbf{0.12} & 0.17 \\
    \textbf{$4^{th}$ t.p. max} &  0.78 & 0.65 & \textbf{0.27} & 0.31 \\
    \textbf{$5^{th}$ t.p. max} &  0.49 & 0.53 & 0.25 & \textbf{0.12} \\
        \bottomrule
    \end{tabularx}
\end{table}
It can be observed that, across all selected metrics, our approach consistently outperforms the state-of-the-art methods. 
As our method is predictive, it effectively anticipates curvature changes and thus outperforms the reactive methods presented in \cite{offset_point_control_article}.
In particular, our approach achieves better performance even when compared to the predictive method presented in~\cite{lukassek_model_2020}.
Moreover, our approach explicitly accounts for sideslip, which significantly reduces tracking errors in curved segments. 
This improvement is reflected in a reduction of approximately $\SI{56}{\%}$ in median tracking error compared to the approach developed in~\cite{lukassek_model_2020}.
Regarding the transient phases, we can observe that predictive methods (the~\ac{MPC} approach presented in~\cite{lukassek_model_2020} and ours) perform better than reactive methods. Furthermore, our approach outperforms the~\ac{MPC} approach presented in~\cite{lukassek_model_2020} in the first, second, and fifth transition phases. 
However, we also observe that the~\ac{MPC} approach presented in~\cite{lukassek_model_2020} performs slightly better in the third and fourth transient phases. 
Finally, we recall that the method presented in \cite{lukassek_model_2020} can only be used for implements mounted at the front of the robot, while ours is generic and works regardless of the implement's position.

\review{
Although no methodological contribution is proposed regarding the sliding observer, an
ablation study was conducted to evaluate its impact. The results reported in
\autoref{tab:observer_ablation} highlight the necessity of using a sliding observer,
particularly during the circular sections $C1$, $C2$, and $C3$.
The median traking error of the implement point is significantly higher when the sliding observer is not used. More specifically, during the circular section $C1$, the use of the sliding observer reduce the tracking
error by $\SI{50}{\%}$. Similarly, increases of $\SI{64}{\%}$ are
observed for sections $C2$ and $C3$. 
As such,  the sliding observer naturally plays a critical role in improving tracking performance, especially during curved paths.}
\begin{table}[thbp]
    \caption{\review{Sliding observer ablation analysis. The median tracking error of the implement point during the circular sections $C1$, $C2$, and $C3$ is reported for the proposed predictive controller with and without the sliding observer
    ($\beta^F = 0$ and $\beta^R = 0$). All values are expressed in meters. ``S.O.''
    stands for Sliding Observer.}}
    \label{tab:observer_ablation}
    \centering
    \begin{tabularx}{\linewidth}{p{2.75cm}XX} 
        \toprule
    \textbf{Implement tracking error}   & \textbf{Predictive (Ours) with S.O.} & \textbf{Predictive (Ours) without S.O. } \\
        \midrule
     \textbf{Median during $C1$} &\textbf{0.03} &  0.06 \\
     \textbf{Median during $C2$} &\textbf{0.05} &  0.14  \\
     \textbf{Median during $C3$} &\textbf{0.04}  &  0.11  \\
        \bottomrule
    \end{tabularx}
\end{table}
\subsection{Agricultural Scenario}
To validate the effectiveness and practical applicability of the proposed control strategy, this section presents results obtained from a representative agricultural scenario. 
\review{In contrast to the previous experiment, the implements were lowered to establish contact with the ground, thereby demonstrating the robustness of the proposed method to implement–soil interaction, during which additional slip is induced due to contact interactions between soil and implement.} 
This scenario is designed to reflect realistic field operations where precise implement control is critical.
One of the common scenarios encountered in agricultural operations involves maneuvers due to the presence of obstacles, such as boulders.
Such obstacles lead to locally complex trajectories, as initially straight paths must be modified to ensure safe avoidance. 
The accuracy of the path-following during such phases is critical to safely avoid the obstacle without hindering agricultural operations. 
This local deviation introduces a form of disturbance during path tracking, challenging the robustness of the control strategy.

\begin{figure}[h]
      \centering
       \includegraphics[width=\linewidth]{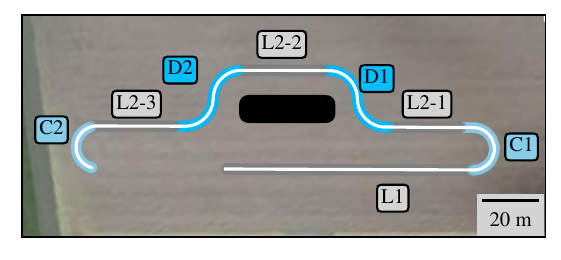}
      \caption{
    Reference path used in the agricultural scenario experiments. 
    The trajectory begins with a straight line (L1), followed by a U-turn (C1). 
    The second segment (L2) is modified to bypass a large natural obstacle, represented by the black box.
    It consists of three straight lines (L2-1, L2-2, L2-3) connected by two deviations (D1, D2), and ends with a U-turn (C2) returning the robot to its initial direction.
      }
      \label{fig:traj_agri_scenario}
\end{figure}

In our case, the reference path illustrated in~\autoref{fig:traj_agri_scenario} consists of two main segments: the first segment is an initial straight line, and the second one is a straight line that is modified to avoid an obstacle. More precisely, it is composed of a straight line $L1$, followed by a U-turn $C1$, then a straight line $L2\text{-}1$, a first deviation $D1$ to avoid the obstacle, a straight line $L2\text{-}2$ to bypass the obstacle, a second deviation $D2$ to return to the line $L2\text{-}3$, and finally a U-turn $C2$ bringing back the robot to the starting direction.

\subsubsection{Agricultural Implements Description}
\begin{figure}[htbp]
    \centering
    \includegraphics[width=\linewidth]{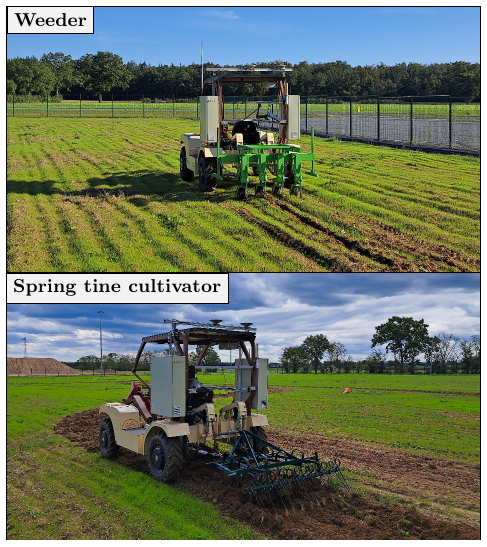}
    \caption{Agricultural implements used in the experiments: a weeder (top) and a spring tine cultivator (bottom). The spring tine cultivator works more deeply than the weeder and is subjected to greater forces.}
    \label{fig:agri_scenario}
\end{figure}
The experiments focus on two commonly used soil maintenance implements, illustrated in~\autoref{fig:agri_scenario}: a weeder and a spring tine cultivator. 
Both implements have long lever arms and are mounted at the rear of the machine, which strongly influence path-following performance and can damage crops if not properly controlled. While the weeder performs surface-level weed removal, the spring tine cultivator penetrates deeper into the soil, thus influencing the overall robot's dynamics.
The control point is located at the rear and the rightmost edge of the implement, positioned at $I_s=\SI{-2}{\meter}, I_y = \SI{-0.5}{\meter}$ relative to the center of the rear axle. 
Such a placement enables a better control of the implement’s spatial footprint. 
The parameters used for the algorithms are the same ones as listed in~\autoref{tab:parameters_expe2} for the Rear-Right case.
Since the implement is mounted at the rear, we only compare our approach with the methods developed in~\cite{offset_point_control_article}. The method proposed in~\cite{lukassek_model_2020}, designed for front-mounted implements, fails to handle rear-mounted configurations and diverges in this case, as previously discussed.

\subsubsection{Results Analysis}
\begin{figure*}[htbp]
  \centering
  \includegraphics[width=\linewidth]{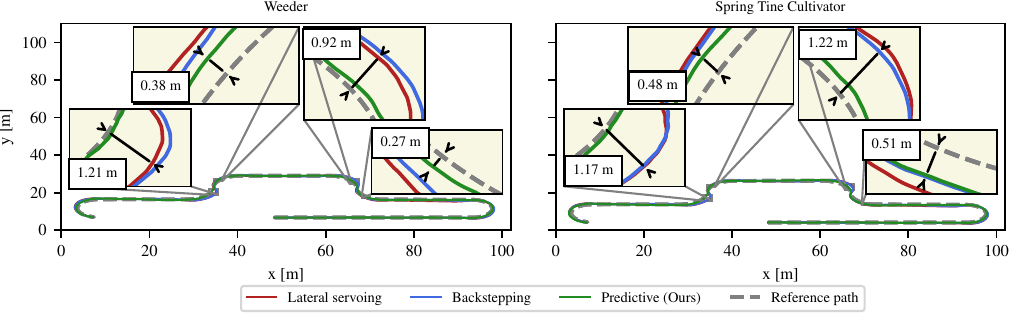}
  \caption{Path-following results for the agricultural scenario using the methods from~\cite{offset_point_control_article} and our proposed approach. 
  The zoomed-in areas highlight the worst tracking errors of the methods in the deviations $D1$ and $D2$, showing that the reactive methods from \cite{offset_point_control_article} lead to a tracking error over \SI{1}{\m}, while our method halves these errors. Note that the trajectory is physically infeasible due to the lever-arm effect of the implement and considerable changes in curvatures, making zero-error tracking unattainable.}
  \label{fig:experiments_agri_scenario_qualitative}
\end{figure*}
We assess the ability of each method to follow the defined reference path and evaluate the performance of each approach to keep the implement close to the desired path during the deviation segments $D1$ and $D2$.
~\autoref{fig:experiments_agri_scenario_qualitative} illustrates the reference path together with the trajectories followed by the implement under each control strategy. Along the straight segments, all methods successfully converge, and the implement closely follows the desired path. A closer look at the deviation segments shows different convergence behaviors in $D1$ and $D2$.

The zoomed-in views around $D1$ and $D2$ show that the reactive methods proposed in~\cite{offset_point_control_article} result in large tracking errors. 
Using the weeder, tracking errors reach up to $\SI{0.92}{\meter}$ in $D1$ and $\SI{1.21}{\meter}$ in $D2$. 
With the spring tine cultivator, tracking errors are even larger, reaching $\SI{1.22}{\meter}$ for $D1$ and $\SI{1.17}{\meter}$ for $D2$. 
Such tracking errors may result in potential collisions with obstacles.
In contrast, our approach remains significantly closer to the reference path: 
with the weeder, the maximum tracking error is limited to $\SI{0.27}{\meter}$ during $D1$ and $\SI{0.38}{\meter}$ during $D2$. With the spring tine cultivator, the tracking errors are up to $\SI{0.51}{\meter}$ for $D1$ and $\SI{0.41}{\meter}$ for $D2$. Note that the reference trajectory is not feasible due to the lever arm effect of the rear implement and large curvature changes, and thus zero-error tracking is not possible.

These observations are summarized in \autoref{tab:quantitative_comparison_agri_scenario}. 
Overall, our approach shows a median improvement of $24\%$ relative to the backstepping method and $41\%$ relative to the lateral servoing.
During the deviation sections, our method significantly outperforms the best alternatives from~\cite{offset_point_control_article}. With the weeder, it reduces tracking errors by \SI{70}{\%} in $D1$ and \SI{69}{\%} in $D2$. 
With the spring tine cultivator, it achieves improvements of \SI{58}{\%} in $D1$ and \SI{55}{\%} in $D2$. 
In all cases, the proposed approach maintains the trajectory closer to the reference path during avoidance maneuvers, offering safer obstacle avoidance while preserving task quality.

\begin{table}[thbp]
    \caption{Quantitative analysis of our method and the approaches of \cite{offset_point_control_article}, including nominal cases on the whole trajectory with median and interquartile range, and edge cases during the deviation segments $D1$ and $D2$.
    The units are in meter ($\SI{}{\m}$).}
    \label{tab:quantitative_comparison_agri_scenario}
    \centering
    \begin{tabularx}{\linewidth}{p{2cm}XXX} 
        \toprule
    \textbf{Implement tracking error}   & \textbf{Lateral servoing}~\cite{offset_point_control_article} & \textbf{Back stepping}~\cite{offset_point_control_article} & \textbf{\review{Predictive} (Ours)}\\
    \multicolumn{4}{c}{\textbf{\textit{Weeder}}}\\
        \midrule
    \textbf{Median} & 0.22 & 0.17 & \textbf{0.13} \\
    \textbf{IQR} & 0.49 & 0.40 & \textbf{0.16} \\
        \midrule
     \textbf{Max during $D1$} &  0.92 & 1.03 & \textbf{0.27}  \\
     \textbf{Max during $D2$} &  1.21 & 1.26 &  \textbf{0.38}  \\
    \multicolumn{4}{c}{\textbf{\textit{Spring Tine Cultivator}}}\\
        \midrule
    \textbf{Median} & 0.22 & 0.18 & \textbf{0.14} \\
    \textbf{IQR} & 0.48 & 0.38 & \textbf{0.23} \\
        \midrule
     \textbf{Max during $D1$} & 1.34 & 1.22 & \textbf{0.51}  \\
     \textbf{Max during $D2$} & 1.20  & 1.17 & \textbf{0.41}  \\
        \bottomrule
    \end{tabularx}
\end{table}

Finally, we analyze how the type of implement affects tracking quality: in the case of the spring tine cultivator, the mean tracking error of our method during segments $D1$ and $D2$ is about $\SI{44}{\%}$ higher than with the weeder. This difference can be attributed to the operating depth of the cultivator, which generates stronger soil interaction forces, leading to more pronounced sliding and additional dynamical effects.


\section{DISCUSSION}\label{sec:discussion}
\review{

In this paper, we consider the control of rigidly linked agricultural implements and outline practical challenges related to real-world deployment.

First, this study confirms that tracking a rigid offset point fundamentally differs from classical robot-centered path-following, from both modeling and control perspectives. 
The position of the implement relative to the rear axle and the associated lever-arm effect strongly influence the resulting trajectories. 
Consequently, reference paths that are trivial for the vehicle’s center may become physically infeasible for the implement, particularly in the presence of abrupt curvature variations. 
This observation emphasizes the necessity of developing control strategies that do not target  perfect tracking, often unattainable in such configurations, but rather aim to minimize unavoidable deviations induced by the system’s kinematic constraints. 
This motivation led to the development of dedicated offset-point controllers, such as those proposed in \cite{offset_point_control_article}.
However, experimental results reveal that purely reactive controllers, even when carefully tuned, perform poorly during transition phases, since curvature transitions systematically introduce disturbances that reactive methods cannot inherently anticipate. 

As such, predictive reasoning is essential to achieve high tracking accuracy. 
This behavior is particularly beneficial in configurations involving rear-mounted implements, as illustrated in the right-hand case of \autoref{fig:forme_de_la_convergence_de_eI}, where the natural convergence dynamics are counterintuitive: the implement initially has to increase its deviation from the reference path before converging. The proposed predictive strategy correctly captures this behavior, whereas classical \ac{MPC} formulations lacking explicit convergence shaping (e.g., \cite{lukassek_model_2020}) fail to converge.

Moreover, the proposed method relies on a closed-form predictive formulation, in contrast with state-of-the-art predictive approaches that require numerical solvers and may suffer from convergence to local minima and non-constant computational time. Experimental results also highlight that sliding effects constitute a source of tracking error, especially for deep-working implements such as spring tine cultivators. Incorporating a sideslip observer substantially improves robustness, especially on  curved segments. Neglecting sliding effects, as in \cite{offset_point_control_article, lukassek_model_2020}, leads to systematic biases on curved paths and increased tracking errors.

It should be noted that the proposed approach has been validated on an Ackermann vehicle platform.
Nevertheless, it can be extended to skid-steering platforms by redefining the control input as
\begin{equation}
    u = \frac{v \tan \delta}{L},
    \label{eq:ackerman_to_skid}
\end{equation}
where $u$ denotes the vehicle angular velocity.
The first stage of the method, which consists of determining the desired angular deviation, remains unchanged, as its formulation is independent of the vehicle platform. In the absence of lateral sliding, the desired angular deviation can still be computed using \autoref{eq:Step01}.
Only the second stage of the method needs to account for the redefinition of the control input introduced in \autoref{eq:ackerman_to_skid}. In the presence of significant sliding effects, dedicated sideslip observers specifically designed for skid-steering vehicles would be required. However, the proposed controller would preserve its cascaded structure: first, the optimal desired angular deviation is computed according to the criterion defined in \autoref{eq:optimisatiocriterion}, and second, the corresponding angular velocity command is derived.
Consequently, the extension of the proposed approach to skid-steering platforms can be achieved by relying on the developments presented in \autoref{sec:theory}, together with the change of control variable defined in \autoref{eq:ackerman_to_skid}.

Finally, the current approach relies on the low-curvature assumption.
While such an assumption is valid in typical agricultural scenarios, where path curvature is small and turning radii are large, it may no longer hold for highly curved reference paths or for vehicles capable of small turning radii, such as skid-steered platforms or compact Ackermann vehicles.
In such cases, deriving a closed-form predictive solution becomes significantly more complex.
Furthermore, although the method’s singularities are not encountered in practice under the low-curvature assumption, a detailed analysis of the robot's behavior in their vicinity would be required if this assumption were relaxed.

These limitations motivate several directions for future work.
\rereview{A first avenue consists of conducting additional real world experiments to better assess robustness and repeatability, as the current validation was limited to a small number of field trials. Such experiments remain challenging in field robotics, where deployments are costly and subject to variable environmental conditions that complicate controlled evaluation.}
A second avenue consists of extending the proposed framework to skid-steered platforms, for which the low-curvature assumption may no longer hold. 
In addition, a study of the trade-offs between trajectory smoothness, tracking performance, and control effort will also be conducted.
Another direction concerns active implements, for which the control point is no longer fixed and introduces additional degrees of freedom. 
Such configurations encountered in vineyard operations with inter-vine blades require additional modeling and control strategies to fully exploit these additional degrees of freedom.}

\section{CONCLUSION AND FUTURE WORK}

\review{In this paper, we presented a predictive control strategy for path-following of a rigid offset point mounted on an agricultural mobile robot. The proposed approach addresses the need for high-precision implement control, which is a key requirement in agricultural operations where the implement directly interacts with crops. 
By explicitly accounting for lateral slip and anticipating curvature transitions, the method overcomes the inherent limitations of classical reactive control strategies. 
Experimental evaluations conducted with two implements demonstrate significant improvements in tracking accuracy and robustness, particularly during curvature transitions and trajectory disturbances. }
\rereview{The proposed method reduces the median of implement tracking errors by $\SI{24}{\%}$ to $\SI{56}{\%}$, and lowers the peak of the tracking errors during curvature transitions by up to $\SI{70}{\%}$.}
Moreover, the proposed controller is applicable to both front and rear-mounted implements, also demonstrating that rear-mounted configurations necessitate nontrivial control strategies for convergence.  \\
\review{
Future work will extend the proposed framework to skid-steered platforms, without relying on the low-curvature assumption. 
Second, we will explore the extension to active implements by modeling the implement control point as an additional degree of freedom, as encountered in systems such as inter-vine blades implements used in viticulture. 
This extension allows the conjoint control of vehicle steering and implement actuation, where the additional degrees of freedom can be used to optimize secondary objectives such as smoothness or implement effort.
}

\printbibliography

\vfill\pagebreak

\end{document}